\documentclass[acmsmall,nonacm,authorversion]{acmart}

\AtBeginDocument{  \providecommand\BibTeX{{    \normalfont B\kern-0.5em{\scshape i\kern-0.25em b}\kern-0.8em\TeX}}}

\usepackage{amsmath}
\usepackage{enumitem}
\usepackage{nicefrac}
\usepackage{wrapfig}
\usepackage{multirow}
\usepackage{xspace}
\usepackage{makecell}
\usepackage{longtable}

\definecolor{Red}{rgb}{1,0,0}
\definecolor{Green}{rgb}{0,0.69,0}
\definecolor{Blue}{rgb}{0,0,1}
\definecolor{LightBlue}{rgb}{0,0.5,1}
\definecolor{veryLightBlue}{rgb}{0.85,0.98,1}
\definecolor{veryLightGreen}{rgb}{0.6,1,0.6}
\definecolor{Skin}{rgb}{1,0.71,0.69}
\definecolor{Grey}{rgb}{0.5,0.5,0.5}
\definecolor{LightGrey}{rgb}{0.6,0.6,0.6}
\definecolor{VeryLightGrey}{RGB}{219, 219, 219}
\definecolor{Black}{rgb}{0,0,0}
\definecolor{White}{rgb}{1,1,1}
\definecolor{brickred}{rgb}{0.8, 0.25, 0.33}
\definecolor{burntOrange}{RGB}{255,122,20}
\definecolor{navy}{RGB}{80, 74, 255}
\definecolor{teal}{RGB}{0, 123, 159}
\definecolor{aquamarine}{RGB}{51, 153, 255}
\definecolor{saffron}{RGB}{227, 170, 0}
\definecolor{purplePink}{RGB}{160, 89, 107}
\definecolor{xanadu}{RGB}{126, 145, 129}

\newcommand{\gender}{\texttt{gender}\ }
\newcommand{\finaleval}{\texttt{Reliable Evaluation}\ }

\newcommand{\ie}{\textit{i}.\textit{e}.\ }

\newcommand{\rulesep}{\unskip\ \vrule\ }

\usepackage{soul}
\usepackage{caption}
\usepackage{subcaption}
\usepackage{tabto}

\begin{document}

\title{Towards Reliable Assessments of Demographic Disparities in Multi-Label Image Classifiers}

\author{Melissa Hall}
\affiliation{ \institution{Meta AI}
 \country{USA}
}
\author{Bobbie Chern}
\affiliation{ \institution{Meta AI}
 \country{USA}
}
\author{Laura Gustafson}
\affiliation{ \institution{Meta AI}
 \country{USA}
}
\author{Denisse Ventura}
\affiliation{ \institution{Meta AI}
 \country{USA}
}
\author{Harshad Kulkarni}
\affiliation{ \institution{Meta AI}
 \country{United Kingdom}
}
\author{Candace Ross}
\authornote{Equal contribution in research and engineering leadership.}
\affiliation{ \institution{Meta AI}
 \country{USA}
}
\author{Nicolas Usunier}
\authornotemark[1]
\affiliation{ \institution{Meta AI}
 \country{France}
}

\makeatletter
\let\@authorsaddresses\@empty
\makeatother

\renewcommand{\shortauthors}{Hall, et al.}

\begin{abstract}
Disaggregated performance metrics across demographic groups are a hallmark of fairness assessments in computer vision.
These metrics successfully incentivized performance improvements on person-centric tasks such as face analysis and are used to understand risks of modern models.
However, there is a lack of discussion on the vulnerabilities of these measurements for more complex computer vision tasks.
In this paper, we consider multi-label image classification and, specifically, object categorization tasks.
First, we highlight design choices and trade-offs for measurement that involve more nuance than discussed in prior computer vision literature.
These challenges are related to the necessary scale of data, definition of groups for images, choice of metric, and dataset imbalances.
Next, through two case studies using modern vision models, we demonstrate that naive implementations of these assessments are brittle.
We identify several design choices that look merely like implementation details but significantly impact the conclusions of assessments, both in terms of \textit{magnitude} and  \textit{direction} (on which group the classifiers work best) of disparities.
Based on ablation studies, we propose some recommendations to increase the reliability of these assessments.
Finally, through a qualitative analysis we find that concepts with large disparities tend to have varying definitions and representations between groups, with inconsistencies across datasets and annotators.
While this result suggests avenues for mitigation through more consistent data collection, it also highlights that ambiguous label definitions remain a challenge when performing model assessments.
Vision models are expanding and becoming more ubiquitous; it is even more important that our disparity assessments accurately reflect the true performance of models.

\end{abstract}

\maketitle

\begin{figure}
         \includegraphics[width=\textwidth]{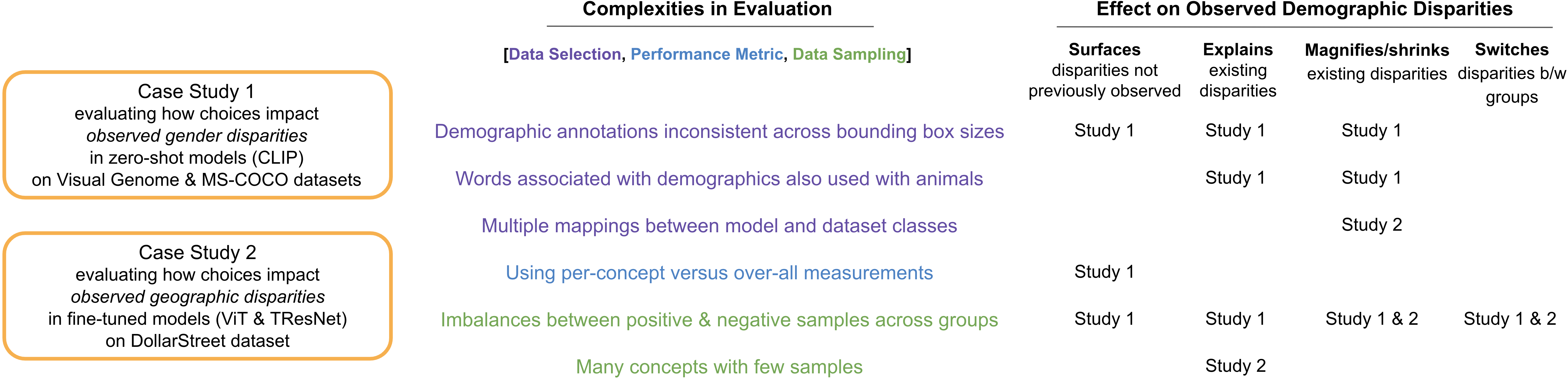}
     \caption{
     In multi-label image classification, we find that the reliability of assessments for demographic disparities are significantly impacted by different evaluation choices.
           First, in Section \ref{sec:choices}, we delineate the main choices and trade-offs of evaluation pertaining to \textit{Data Selection}, \textit{Performance Metric}, and \textit{Data Sampling}.
      Next, we perform two case studies in Sections \ref{sec:case1} \& \ref{sec:case2} using modern object recognition models.
      We demonstrate that the \textbf{complexities in evaluation} related to the aforementioned choices have an \textbf{effect on observed demographic disparities}: they can \textit{surface} new disparities, \textit{explain} existing disparities, \textit{magnify or shrink} existing disparities, and \textit{switch} disparities between groups.
    }
    \label{fig:main_fig}
\end{figure}

\section{Introduction}
\label{sec:introduction}
As computer vision systems become ubiquitous, there is a growing collection of work focused on assessing ``who these systems work for'' \citep{Mohamed_2020,10.1145/3531146.3533083}.
This question is often explored through disaggregated measurements \citep{barocas2021designing}, which compare performance across groups of people affected by these systems.
Audits of face analysis systems are an example of disaggregated measurements that highlight problematic error discrepancies between groups and enable the monitoring of improvements \citep{buolamwini2018gender,raji2019actionable}.
As vision models grow in capabilities, fairness measurements of feature extractors and zero-shot learners trained on billions of images similarly rely on disaggregated measurements \citep{goyalfairness2022,Singh_2022_CVPR}.
A recent subset of analyses focus on \emph{per-label} disaggregated measurements for \emph{multi-label} object recognition tasks \citep{zhao2017men,https://doi.org/10.48550/arxiv.2102.12594,radford2021learning,agarwal2021evaluating,https://doi.org/10.48550/arxiv.2301.11100}.

Compared to fairness assessments of object recognition tasks that use overall performance metrics \citep{devries2019does,goyal2022vision,rojas2022dollar}, the finer granularity offered by per-label measurements reveals discrepancies that may be hidden by overall metrics.
For example, an \textit{overall measurement} may show that a model is equally good at identifying objects when images contain only men or only women.
A \textit{per-label measurement}, on the other hand, can provide finegrained detail such as finding that the model has struggled to classify neckties when men are not in the image and equally struggled for handbags when women are not in the image.
These measurements also explicitly link error discrepancies with specific labels, making them a useful tool to better understand model errors.

Disaggregated measurements for multi-label classification are challenging because they are, by nature, \emph{imperfect} representations of model errors.
They are summary statistics that give only a high-level view of error patterns and are subject to confounding factors, including ambiguities in the determination of groups and correlations between labels and groups.
Choices made for measurement directly impact these confounders and the reliability of the assessment.
While these effects have been discussed in isolation in the machine learning and algorithmic fairness community \citep{barocas2021designing}, per-concept, multi-label object recognition involves complexities that deserve specific attention.
Multi-label tasks require an examination of metrics beyond binary classification, which has received the most interest so far \citep{chouldechova2017fair,kleinberg2018inherent,golz2019paradoxes}.
Furthermore, there is no high-fidelity benchmark for object recognition that contains demographic attributes of people and captures many concepts co-occurring in images.
The main datasets use model-provided annotations of objects that can have their own biases \citep{miap_aies} or focus on location-based disparities with few labels per image  \citep{devries2019does,rojas2022dollar}.

The contributions of this paper are three-fold:
\begin{itemize}[noitemsep,nolistsep]
    \item First, in Section \ref{sec:choices} \textbf{we highlight design choices that are unique to per-concept disaggregated measurements in multi-label object recognition tasks}, including the data source and specification of groups, trade-offs when using multi-label data, and the interplay between metrics and sampling.
    \item Second, in Sections \ref{sec:case1} \& \ref{sec:case2} we perform two case studies that demonstrate how design choices affect observed performance disparities.
        \textbf{We find that evaluation choices affect the magnitude and direction of observed disparities}.
    For example, disparities in average precision between groups are dependent on the ratio of positive and negative samples for each group, and different mappings between dataset and model labels can magnify observed disparities between groups. We propose recommendations to improve assessments.
    \item Finally, in Section \ref{sec:root_cause} we qualitatively analyze disparities in the more reliable evaluation protocols of our case studies.
    \textbf{We observe that concepts with large disparities often have ambiguous definitions, inconsistent annotations, or domain shifts}.
    The variation of concept representations between groups suggests avenues for disparity mitigations while also posing a remaining challenge for performing reliable evaluations.
\end{itemize}

\vspace{1mm}

\noindent \textit{\textbf{Limitations.}} Our work emphasizes the importance of carefully designing measurements disaggregated by group and concept.
It also highlights trade-offs when evaluating disparities in object recognition tasks.
While we believe our work will help practitioners better control sources of observed disparities that are not due to group-dependent model errors, we reiterate that disaggregated measurements are by nature imperfect.
First, exactly controlling for within-group label correlations is likely infeasible.
Second, the existing specifications of demographic groups are far from capturing the complexity and nuances of these constructions \citep{benthall2019racial,hanna2020towards,hazirbas2021towards,dodik2022sex,Tomasev_2021}.
For example, the operationalization of gender through a binary variable and proxy labels as we do for the Visual Genome dataset in Section \ref{sec:case1} is only a best effort to obtain high-level correlations between errors and should not be seen as a ``correct'' measurement of gender-based disparities (if that exists at all).
We discuss these risks more in Section \ref{sec:choices} and Section \ref{sec:case1}.
Finally, while we discuss the interplay between metrics and sampling (see Section \ref{sec:choices}), \emph{which} metric to choose is still left as an open question -- which is all the more important considering incompatibility between metrics \citep{kleinberg2018inherent,chouldechova2017fair}.

These limitations, in turn, affect the interpretation and use of these measurements.
In this work, we use disaggregated measurements to guide a qualitative assessment of model errors.
Similar to \citet{goyalfairness2022}, they can also be used to measure progress.
However, remaining unknowns and risks related to the definition of groups and choice of metric make these measurements insufficient to certify fairness or provide fairness constraints.

\section{Related work}
\label{sec:related_work}
Measuring and understanding bias in vision models typically focuses on either training data \cite{zhao2017men,wang2019balanced,wang2022revise} or the assessment of the model after training. We only discuss the latter in this section.

In the last decade, audits of AI systems based on disaggregated measurements showed higher errors for societally disadvantaged demographic groups \citep{grother2014face,angwin2016machine,https://doi.org/10.48550/arxiv.1711.11443,buolamwini2018gender}.
For recent image recognition and classification models \citep{rojas2022dollar,Singh_2022_CVPR,goyal2022vision,radford2021learning}, fairness assessments crystallized primarily around two issues: disparities in object recognition performance \citep{shankar2017no,devries2019does,Singh_2022_CVPR,rojas2022dollar,zhao2017men} and identifying harmful or stereotypical associations \citep{radford2021learning,goyalfairness2022,zhao2017men,https://doi.org/10.48550/arxiv.2301.11100}.
A critical challenge in disaggregated analyses of object recognition models is the availability of datasets with clear demographic annotations.
Data collection has mostly focused on annotations of demographic groups for images of faces and people ~\citep{hazirbas2021towards,schumann2021step,karkkainenfairface,buolamwini2018gender}, rather than explicit object annotations.
While some datasets do contain object annotations in addition to geographic or income information \citep{dollarstreet,rojas2022dollar,ramaswamy2022geode}, they contain few labels per image (no more than five) and are more suited for \emph{single-label} evaluations.
Open Images MIAP ~\citep{miap_aies} is an object recognition dataset with explicit gender and age annotations, but its object labels are supplied by models \citep{miap_aies,DBLP:journals/corr/abs-1811-00982} rather than humans and may contain model-induced biases.
Prior work \citep{zhao2017men,wang2019balanced,https://doi.org/10.48550/arxiv.2102.12594,https://doi.org/10.48550/arxiv.2301.11100} also adapts general purpose object recognition datasets by constructing proxies for gender groups based on annotations of gender-related terms.
Our contribution includes a review of the design choices in these applications, such as how groups are deduced and the experimental setup, and we propose steps to make disparity measurements based on these datasets more reliable and clarify their limitations.

The choice of metrics has been debated in the context of bias in recidivism prediction tools \citep{angwin2016machine,flores2016false,dieterich2016compas}, as well as their incompatibilities \citep{chouldechova2017fair,kleinberg2018inherent}.
Metrics for fairness assessments in computer vision receive less focus:  precision/recall or true/false positive rates \citep{buolamwini2018gender} and hit-rate \citep{devries2019does,goyalfairness2022,Singh_2022_CVPR} are used without critical discussion.
Our work brings metrics to the forefront.
In particular, the tuning of decision thresholds poses a difficulty for multi-label predictions and suggests the use of ranking metrics to avoid the issue.
However, we show that for the commonly used average precision metric, group/label correlations have a major, undesirable impact on observed disparities.

Previous work studying disparities within vision models focuses on different representations of a \emph{single} concept definition, noting potential factors of variation across languages \cite{devries2019does} and visual cues \cite{https://doi.org/10.48550/arxiv.2211.01866}.
Our qualitative analysis highlights how - beyond different translations - a concept's varied meaning between demographic groups can affect model performance.
For example, in the Visual Genome dataset, the concept ``base'' is often a baseball field base in images containing people annotated as men and a structural base of an object for images containing annotations of women.

By investigating the malleability of concept definitions, our work is in line with recent works studying the breadth of concept meanings (for example, supplementing ImageNet with more annotations \citep{DBLP:journals/corr/abs-2101-05022, DBLP:journals/corr/abs-2006-07159, pmlr-v119-shankar20c}), as well as examinations of classifications as evolving artifacts of the relationship between perception and inherent meaning \citep{doi:10.1177/20539517211035955,10.7551/mitpress/6352.001.0001,border,bender-koller-2020-climbing}.

\section{Choices Matter in Measurement}
\label{sec:choices}
\def\negtab{\hspace{-2em}}
\def\vs{\textit{vs}\ }
\def\dash{--\ }

\begin{table}[!tb]
    \centering

    \scalebox{0.85}{
    \begin{tabular}{p{14em}p{14.5em}p{14em}}
    \centering \bf Data Selection
    &
    \centering \bf Performance Metric
    &
    \multicolumn{1}{c}{\bf{Data Sampling}}
    \\
    \multicolumn{3}{c}{\it Design Choices}\\

    \begin{enumerate}[wide, labelwidth=!, labelindent=1pt, leftmargin=*]
        \item[A)] data source
        \item[B)] operationalization of groups
        \item[C)] data \& model class mapping
    \end{enumerate}

    &
    \begin{enumerate}[wide, labelwidth=!, labelindent=1pt, leftmargin=*]
        \item[D)] precision/recall or TP/FP rates
        \item[E)] ranking \vs classification
    \end{enumerate}

    &
    \begin{enumerate}[wide, labelwidth=!, labelindent=1pt, leftmargin=*]
        \item[F)] filtering rare labels
        \item[G)] controlling for correlations
    \end{enumerate}
    \\

    \multicolumn{3}{c}{\it Trade-Offs}\\

    \begin{enumerate}[wide, labelwidth=!, labelindent=1pt, leftmargin=*]
    \item[A)] annotation cost \textit{vs} \newline quality
    \item[B)] data coverage \vs \newline relevance to group
    \end{enumerate}

    &
    \begin{enumerate}[wide, labelwidth=!, labelindent=1pt, leftmargin=*]
    \item[C)] needs tuning thresholds \vs \newline avoids tuning thresholds

    \item[D)] needs fixing $+/-$ class ratios \vs \newline avoids fixing $+/-$ class ratios
    \end{enumerate}
    &

    \begin{enumerate}[wide, labelwidth=!, labelindent=0pt, leftmargin=*]
    \item[E)] data coverage \textit{vs} \newline statistical significance
    \item[F)] control for confounders \textit{vs} \newline distribution shift
    \end{enumerate}

    \end{tabular}
    }

    \caption{Main design choices and trade-offs for disparity analyses of multi-label classification tasks.
}
    \label{tab:design_choices}
\end{table}

In this section we discuss important design choices for per-label, disaggregated measurements of object recognition models.
We focus on recognition tasks defined by popular datasets like the Visual Genome \citep{krishna2017visual}, MS-COCO \citep{lin2014microsoft} or Dollar Street \citep{rojas2022dollar}.
These datasets contain object annotations, such as ``tie'' and ``showers,'' and have been used to analyze disparities between gender, geography, and income groups \citep{devries2019does,zhao2017men,https://doi.org/10.48550/arxiv.2301.11100}.
We study cases when disparities in performance measurements may not be due to the model, but rather, idiosyncrasies or biases in \emph{how} we perform the measurement.

We aim to improve the reliability of the assessment so that observed discrepancies between groups represent, as much as possible, actual differences in classification behavior between groups.
We emphasize considerations that are particularly important for per-label disaggregated measurements of object recognition.
Table~\ref{tab:design_choices} summarizes our  exposition of design choices and trade-offs and covers
\textbf{Data Selection}, \textbf{Performance Metric}, and \textbf{Data Sampling}.

We ground our discussion with examples from two case studies investigating the effect of these choices on observed disparities:
In Case Study \#1 we study gender-related disparities in the zero-shot CLIP model \cite{radford2021learning} using the Visual Genome and MS-COCO, and in Case Study \#2, we evaluate geographic disparities between supervised object recognition models using Dollar Street.
We present these studies in Section \ref{sec:case1} and Section \ref{sec:case2} and summarize our learnings in Figure \ref{fig:main_fig}.

\subsection{Data selection}

The choice of data source is tied to the demographic groups that are the target of the evaluation and model task.

\vspace{1mm}

\noindent \textit{\textbf{Choice A: Data source.}}
For object recognition, existing datasets dedicated to disparity analyses address questions related primarily to \emph{where} the image was taken, containing location or household-based income metadata \citep{shankar2017no,rojas2022dollar,ramaswamy2022geode}.
To date, the only dataset that assesses performance of object classification with respect to annotated demographic attributes of \emph{people in the image} contains object labels that are provided by classifiers, rather than humans ~\cite{miap_aies}.
The dataset likely inherits any biases contained in the models used for annotations, making it not ideal for disparity evaluations.
To enable disaggregated measurements of more groups, prior work \citep{zhao2017men,wang2019balanced,https://doi.org/10.48550/arxiv.2102.12594,https://doi.org/10.48550/arxiv.2301.11100} constructs group labels from existing annotations in object recognition datasets that pertain to people.
The main trade-off in the choice of data source relates to the cost of annotations: Compared to adapting existing datasets, collecting a dedicated dataset is expensive, but offers more control over how the groups are operationalized and which images compose the evaluation dataset.
We discuss these further in the following paragraphs.

\vspace{1mm}

\noindent \textit{\textbf{Choice B: Operationalization of groups.}}
In images, group information pertaining to the individuals shown can be based on self-identification \citep{hazirbas2021towards} or ``perception'' \citep{schumann2021step} by an external source.

Inferring group identity from pre-existing annotations that are not focused on assigning demographic information introduces additional challenges.
First, labels used for determining groups are likely not applied consistently by annotators, since their intent is not to explicitly operationalize the perceived group.
For example, annotators may mention demographic groups more often when the situation depicted is unexpected for them, such as noting a ``female doctor'' when a woman is shown and a ``doctor'' when a man is shown.
Thus, the inclusion of the demographic term can be inconsistent and encode annotator biases.
Second, annotators may use words in unexpected ways.
For example, in Case Study \#1 we find that annotators of the Visual Genome often label images of animal pairs as ``mother'' and ``child.''

Furthermore, disaggregated measurements assign group information \emph{to images}, while demographic attributes are characteristics \emph{of individuals}.
This distinction has not received much consideration in prior work
because models have been primarily assessed on images of individual people \citep{buolamwini2018gender,radford2021learning,goyalfairness2022}.
Mapping an image to a single demographic group is more challenging when
multiple people are in the photo.
This introduces a trade-off: filtering images that are difficult to assign groups
means less noise in terms of group labels but reduces coverage and inclusivity of the assessment.
For example, prior work studying gender disparities removes images with reference to multiple groups to allow for distinct groups, even though doing so excludes many images \citep{zhao2017men}.
In Case Study \#1, we also investigate filtering images to ensure annotators had some minimum amount of information present for perceiving groups.

Finally, adapting pre-existing datasets means the only possible demographic groups are those explicitly labeled.
For instance, while using captions to create group proxies might extend to age using references words like ``young'' or ``old'', they do not extend to attributes that annotators do not reference consistently, such as skin tone or ethnicity.

While inferring group annotations from existing datasets provides insights of high-level error patterns of models, they are not sophisticated enough to provide fairness certificates and should not preclude the development of new datasets designed specifically for disparity analyses.

\vspace{1mm}

\noindent \textit{\textbf{Choice C: Data and model class mapping.}}
Evaluation datasets contain explicit object labels for each image.
Like words, labels can have multiple meanings that are implied or modified depending on who is applying them.
Object recognition models may be trained with class labels (concepts) that do not align with evaluation dataset labels, or collected with different specifications.
In these cases, practitioners need to determine a mapping between evaluation dataset labels and model concepts.
This presents a trade-off: creating this mapping means more concepts can included but a mis-match between dataset and model definitions means a less indicative evaluation of intended model performance.
For example, in Case Study \#2 we investigate how variation in mapping of concepts such as ``parking lots'' to ``garage'' versus ``parking meter'' can affect observed disparities.

\subsection{Performance metric}

For per-label disaggregated measurements of multi-label classifiers, we follow a one-vs-all approach: for each label, there is an associated binary classification task determining the presence or absence of the label in an image.
We can then measure per-label disparities using similar metrics as with binary classification.
In the discussion below, the \emph{positive class} refers to the label being present in a given image and \emph{negative class} refers to the label being absent.

\vspace{1mm}

\noindent \textit{\textbf{Choice D: Choosing the family of metrics.}}
The main families of metrics in binary classification are accuracy, precision/recall, and true positive/false positive rate (TPR/FPR) \citep{manning1999foundations}.
TPR and FPR are the proportions of predicted positives within the true positive examples and true negative examples, respectively.
TPR and recall are the same, while precision is the proportion of true positive examples within the predicted positive.

The fundamental difference between precision/recall and TPR/FPR analyses lies in their respective dependence on the proportion of positive examples, called \emph{prevalence}.
Estimates of TPR/FPR remain approximately the same if we resample the data by changing the prevalence, but this is not the case for the precision.
To be concrete, precision is related to TPR and FPR by \citep{davis2006relationship}:
\begin{align}
\mathrm{precision}=\frac{\alpha \mathrm{TPR}}{\alpha \mathrm{TPR}+(1-\alpha) \mathrm{FPR}}.
\end{align}
For instance, when $\alpha$ is small, precision is small unless the ratio TPR/FPR is large.
This is why precision/recall analysis is sometimes preferred to TPR/FPR analysis for imbalanced data \citep{saito2015precision}.

However, when comparing two groups, TPR/FPR discrepancies are not affected by the relative numbers of positive/negative examples in each group, but precision is.
In most vision datasets the proportion of images sampled within each class is a question of dataset construction.
We argue that a dependence on prevalence within each group is undesirable: it creates disparities that do not reflect models' errors but, rather, reflect correlations between label and group.
Consequently, we argue that assessments should either focus on TPR/FPR discrepancies or guarantee the same proportion of positive and negative examples for every group when using precision or recall.

Accuracy is less suitable than precision/recall or TPR/FPR for computing per-label disparities:
it suffers from the same drawback as precision since \begin{align} \mathrm{accuracy}=\alpha \mathrm{TPR}+(1-\alpha)(1-\mathrm{FPR}),
\end{align} while also having a low sensitivity to TPR when $\alpha$ is small (often the case in object classification).

\vspace{1mm}

\noindent \textit{\textbf{Choice E: Ranking \textit{vs} classification metrics.}}
Modern computer vision systems based on neural networks do not directly output binary classifications:
models produce confidence scores for each label, and practitioners select decision thresholds to determine whether a label is present.
The choice of the thresholds is application-dependent \citep{fawcett2006introduction} and requires additional data and rules for models such as CLIP, where the scores do not have a meaningful scale \citep{agarwal2021evaluating}.
Since there is no clear definition of a ``good'' threshold, ranking metrics are often used for fairness assessments.

The common metrics are the area under the ROC curve (AUC-ROC) \citep{fawcett2006introduction} and the average precision (AP).
The ROC curve plots TPR as a function of FPR for classifiers obtained by varying the threshold from minimum to maximium score.
AUC-ROC is the area under this curve.
AP is an approximation of the area under the precision-recall curve \citep{davis2006relationship,su2015relationship}, which computes the mean of the precision at each threshold for which there is an increment in recall.
AP is often preferred over AUC-ROC on imbalanced data for the same reason precision/recall is preferred to FPR/TPR analysis, but it suffers from the same drawback when computing disparities, \ie a heavy dependence on prevalence.
As with precision/recall, we argue that computing disparities using AP requires sampling the same number of positive/negative examples for each group and demonstrate the effects of doing so empirically in Case Study \#1 and Case Study \#2.

\subsection{Data sampling}
The last step in measurement is determining which images and labels to use.

\vspace{1mm}

\noindent \textit{\textbf{Choice F: Filtering rare labels.}} It is not possible to provide statistically reliable estimates of disparities for rare labels. A standard solution is to remove labels with less than $K$ occurrences, where $K$ is chosen as a trade-off between the size of confidence intervals and coverage of data.

\vspace{1mm}

\noindent \textit{\textbf{Choice G: Controlling for uneven label correlations.}} As explained when discussing metrics above, it is desirable to resample, \emph{for each label}, the positive and negative class for each group to ensure the prevalence is the same across groups when using precision-based metrics (including AP).
More generally, confusion matrices are non-uniform, and some labels are more easily confused with each other.
This means, for every metric, each change in label distribution across groups can affect observed disparities.
In multi-label data, label correlations differ per group (for instance, ``suit'' and ``tie'' may co-occur more within images of people who are perceived to be men than people who are perceived to be women), and controlling for the marginal distribution of each label is not enough.
It is difficult to achieve, through selective sampling, a target distribution of labels (including co-occurrences) for all groups in the case of multi-label data.
In this work, we focus solely on controlling positive and negative class ratios across groups, without fixing the true label distribution to ensure the same distribution of co-occurring concepts.

\section{Case Study \#1: Evaluating a Single Zero-Shot Model}
\label{sec:case1}

\begin{figure}[!tb]
     \begin{subfigure}[b]{0.99\textwidth}
         \includegraphics[width=\textwidth]{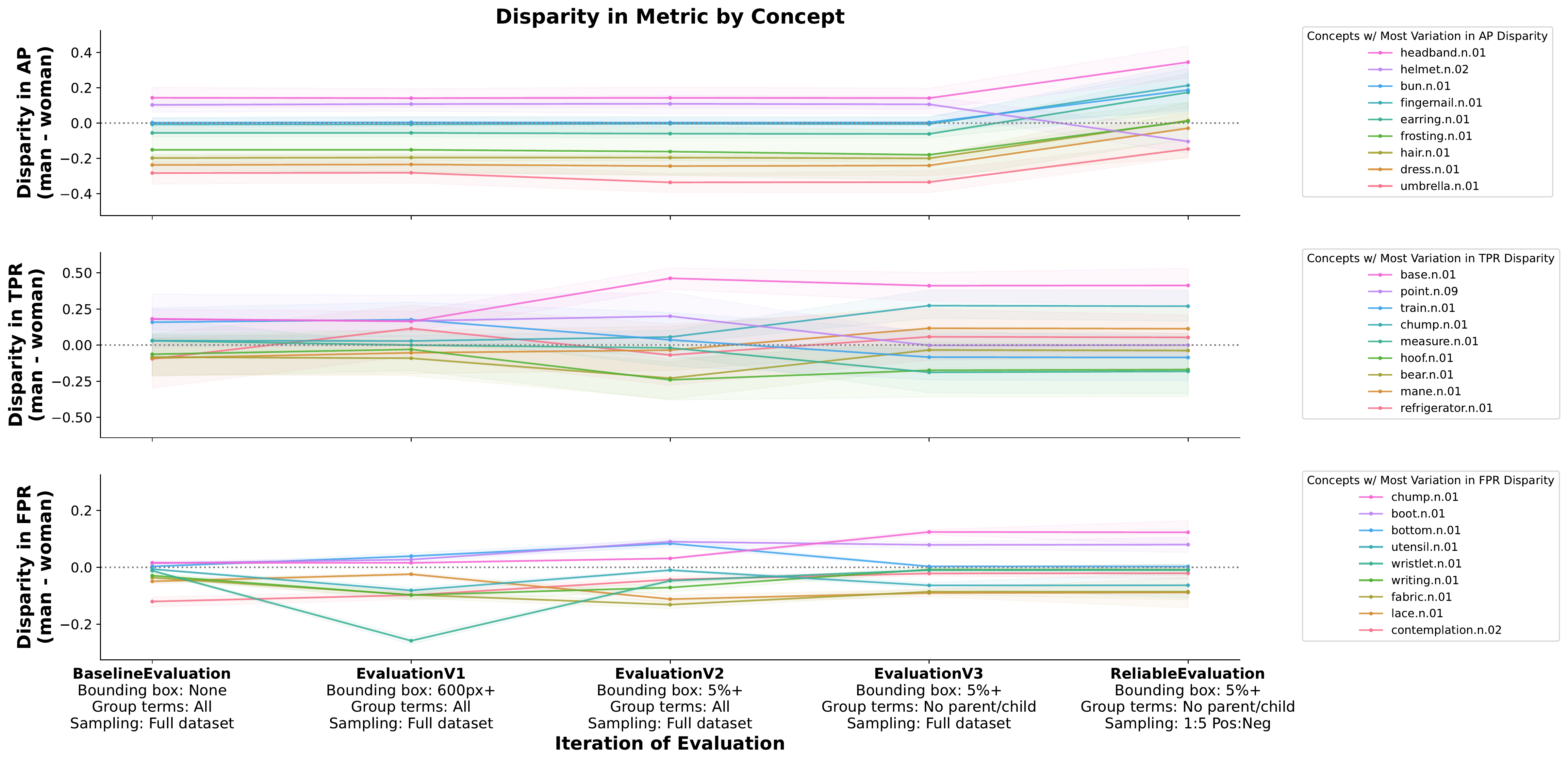}
     \end{subfigure}
     \caption{
     The operationalization of groups and sampling affects the magnitude and direction of per-concept disparities in CLIP in Case Study \#1.
     A positive disparity means that the model has a higher value for the respective metric for the \texttt{man} group than the \texttt{woman} group.
               The shaded error bars are 95\% percentile intervals constructed with 250 bootstraps.
     See Appendix \ref{app:cs1_res} for a per-concept view of the effect of evaluation choices.
     }
     \label{fig:case1_path_eval}
\end{figure}

In this section, we investigate how the design choices discussed in Section \ref{sec:choices} affect observations of demographic disparities in a multi-label classifier.
We begin with a baseline protocol (denoted as \texttt{Baseline Evaluation}) that follows prior work, then adapt the measurement methods for several design choices with the goal of evaluations that better capture model performance.
The intermediate evaluations are denoted as \texttt{Evaluation V[X]} and the last iteration is denoted as the \finaleval method.
While the \finaleval method is by no means the only or best form of evaluation, it addresses some weakness of the \texttt{Baseline Evaluation} (discussed in the Results section below) and allows for an insightful exploration of model performance disparities (discussed in Section \ref{sec:root_cause}).

\subsection{Evaluation set-up}\label{cs1:eval}

We focus our work on the CLIP ViT-B/32 \cite{radford2021learning}, which jointly trains an image
encoder and text encoder to enable zero-shot classification.
While previous works have evaluated CLIP's performance between demographic groups \cite{agarwal2021evaluating,radford2021learning,https://doi.org/10.48550/arxiv.2301.11100,9656762,https://doi.org/10.48550/arxiv.2205.10764}, there has been little discussion of \emph{how} the method of assessment affects observed disparities.
Following \citep{https://doi.org/10.48550/arxiv.2301.11100}, we aim to identify concepts for which CLIP performs disparately between gender-based groups.

\vspace{1mm}

\noindent \textbf{Data selection}

\vspace{1mm}

\noindent \textit{\textbf{Choice A: Data source.}}
As explained in Section \ref{sec:choices}, there are no existing datasets with explicit, human-collected annotations for multiple objects per image and gender-based groups.
We follow \citep{zhao2017men,wang2019balanced,https://doi.org/10.48550/arxiv.2102.12594,https://doi.org/10.48550/arxiv.2301.11100} and use the Visual Genome \cite{krishna2017visual} and MS-COCO \cite{lin2014microsoft} datasets.
The Visual Genome contains 108k images, where each image is annotated with a set of bounding boxes and object labels.
The labels correspond to synsets from WordNet \citep{miller-1994-wordnet}, and each synset is a node representing a singular concept.
COCO has 123k images containing bounding boxes for 80 objects and 5 captions per image.
To prevent overlap, we filter out all images from COCO that are in the Visual Genome.

\vspace{1mm}

\noindent \textit{\textbf{Choice B: Operationalization of groups.}} The Visual Genome and MS-COCO do not contain explicit group annotations. \citep{zhao2017men,wang2019balanced,https://doi.org/10.48550/arxiv.2102.12594,https://doi.org/10.48550/arxiv.2301.11100} use pre-existing object labels and captions to determine binary gender groups for images.
We follow these works to investigate how this methodology can affect observed disparities.
However, we note that definitions of demographic groups that categorize intrinsically continuous personal attributes have the potential for harm, and this is exacerbated when the mapping is performed by an external party \citep{scheuerman2019computers}.
Furthermore, the binarization of gender is exclusive of other genders \citep{https://doi.org/10.48550/arxiv.2205.02526} as well as images that contain multiple groups.
We use the term \gender with groups \texttt{\{man, woman\}} to convey our specific operationalization of groups.

For our first evaluation, \texttt{Evaluation Baseline}, we follow previous work \citep{zhao2017men,wang2019balanced,https://doi.org/10.48550/arxiv.2102.12594,https://doi.org/10.48550/arxiv.2301.11100} and use a list of gender-based terms (see Appendix \ref{app:cs1_terms}) to assign \gender groups to each image.
Images with multiple group labels are removed.
We initially include all bounding boxes containing gender-based terms when determining groups for the Visual Genome\footnote{Because we only use captions for determining groups with COCO, filtering by annotation size is not relevant.}.

We find many of these bounding boxes are small with respect to the overall image size, likely making these annotations less reliable.
With small bounding boxes, information related to gender presentation has low salience and the potential for annotator bias increases.
Therefore, our next two evaluations focus on bounding box sizes. In \texttt{Evaluation V1} we follow \cite{https://doi.org/10.48550/arxiv.2301.11100} and require that bounding boxes are 600+ pixels.
In \texttt{Evaluation V2} we use an alternative filtering schema in place of pixels: images must contain at least one \gender annotation that fills 5+\% of the image and can have \gender annotations smaller than 2\% of the image.
We do not use the latter annotations for assigning groups.
Focusing on relative instead of absolute bounding box sizes is helpful for the Visual Genome where not all images are the same size.

Our next evaluation, \texttt{Evaluation V3}, focuses on
terms associated with gender in images that do not actually contain humans.
For example, large and small animal pairings are frequently tagged with ``mother'' and ``child'' terms,  adding noise to our evaluation.
We remove the gender terms related to this relationship, denoted in Appendix \ref{app:cs1_terms}.

\vspace{1mm}

\noindent \textit{\textbf{Choice C: Data \& model class mapping.} }
The Visual Genome contains human-labeled, free-form object annotations that are mapped to canonicalized synsets using the WordNet \cite{miller-1994-wordnet} hierarchy, such as ``dress.n.01'' or ``bat.n.05.''
To reduce noise of multiple labels referring to the same object and increase our per-concept sample size, we use the synset mappings of each object label.
For MS-COCO, we use object categories provided in the dataset, such as ``person'' or ``bicycle''.

Because of CLIP's zero-shot capability, the Visual Genome and MS-COCO labels do not need to be mapped to CLIP-compatible classes.
Following standard practice, we leverage prompt engineering \cite{radford2021learning,https://doi.org/10.48550/arxiv.2301.11100} to average CLIP predictions across 80 templates for each class, corresponding to phrases like ``a photo of my $\{\}$'' and ``an origami $\{\}$.''
We truncate the synset texts to include only the text corresponding to the object itself and replace underscores with spaces.
\vspace{1mm}

\noindent \textbf{{Performance metric}}

\vspace{1mm}

\noindent \textit{\textbf{Choice D: Choosing the family of metrics.} }
To understand the interplay between metric and sampling, we include both average precision (AP) and true positive rate/false positive rate (TPR/FPR).

\vspace{1mm}

\noindent \textit{\textbf{Choice E: Ranking \textit{vs} classification metrics.} }
Average precision is a ranking metric and does not require a threshold.
TPR and FPR both require thresholding CLIP's output, which is a similarity score between the labels and images.
To determine this threshold, for each evaluation we split the data into a 20\%/80\% validation-test split and pick a threshold for each concept that maximizes the F1-score on the validation split, disallowing labeling everything as negative.

\vspace{1mm}

\noindent \textbf{{Data sampling}}

\vspace{1mm}

\noindent \textit{\textbf{Choice F: Filtering rare labels.} }
Following \cite{https://doi.org/10.48550/arxiv.2301.11100}, in the \texttt{Evaluation Baseline} we remove concepts with less than 50 images per group and images without labels,
yielding 364 concepts for the Visual Genome and 72 concepts MS-COCO.

\noindent \textit{\textbf{Choice G: Controlling for correlations.}} In the \texttt{Evaluation Baseline}, we follow \cite{https://doi.org/10.48550/arxiv.2301.11100,zhao2017men} and do not perform additional sub-sampling.
However, as discussed in Section \ref{sec:choices}, AP has a heavy dependence on prevalence.
In the \texttt{Reliable Evaluation} we build on our previous adaptations and introduce a sampling ratio of 1:5 positive to negative samples for each group to attempt to control for variations in prevalence between groups.
We also require that all objects use this same ratio to additionally control for variations in prevalence among concepts.
This ratio allows us to include all possible concepts while still ensuring a reasonable number of samples.
For each concept, we determine the maximum number of samples such that each group can have the same number of samples and maintain our desired sampling ratio.
For each group, we select positive and negative samples randomly, maintaining the desired sampling ratio and sampling with replacement to perform bootstrapping.
We explain our construction of confidence intervals in Appendix \ref{app:cs1_ci}.

\subsection{Results}

\begin{figure}
     \begin{subfigure}[t]{0.17\textwidth}
         \centering
         \raisebox{5mm}{
         \includegraphics[width=\textwidth]{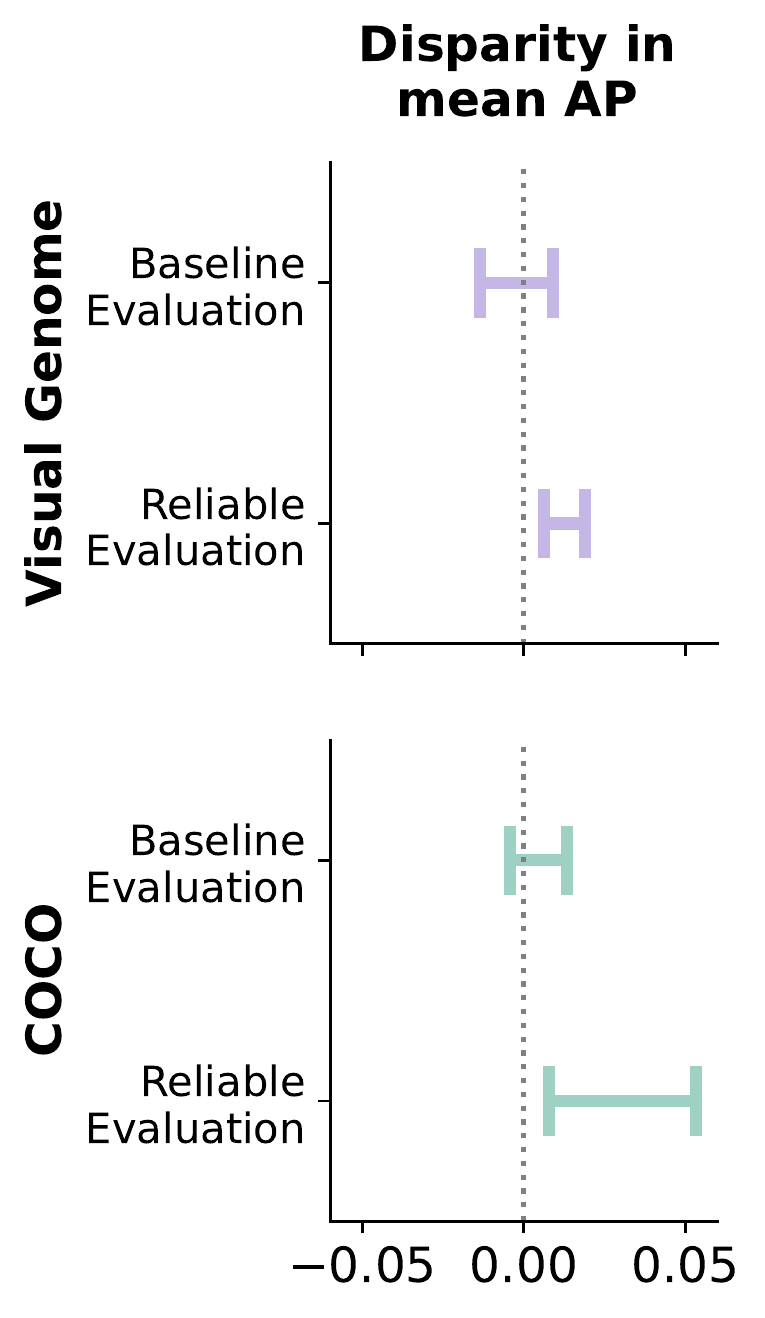}}
     \end{subfigure} \rulesep
     \begin{subfigure}[t]{0.80\textwidth}
         \centering
         \includegraphics[width=\textwidth]{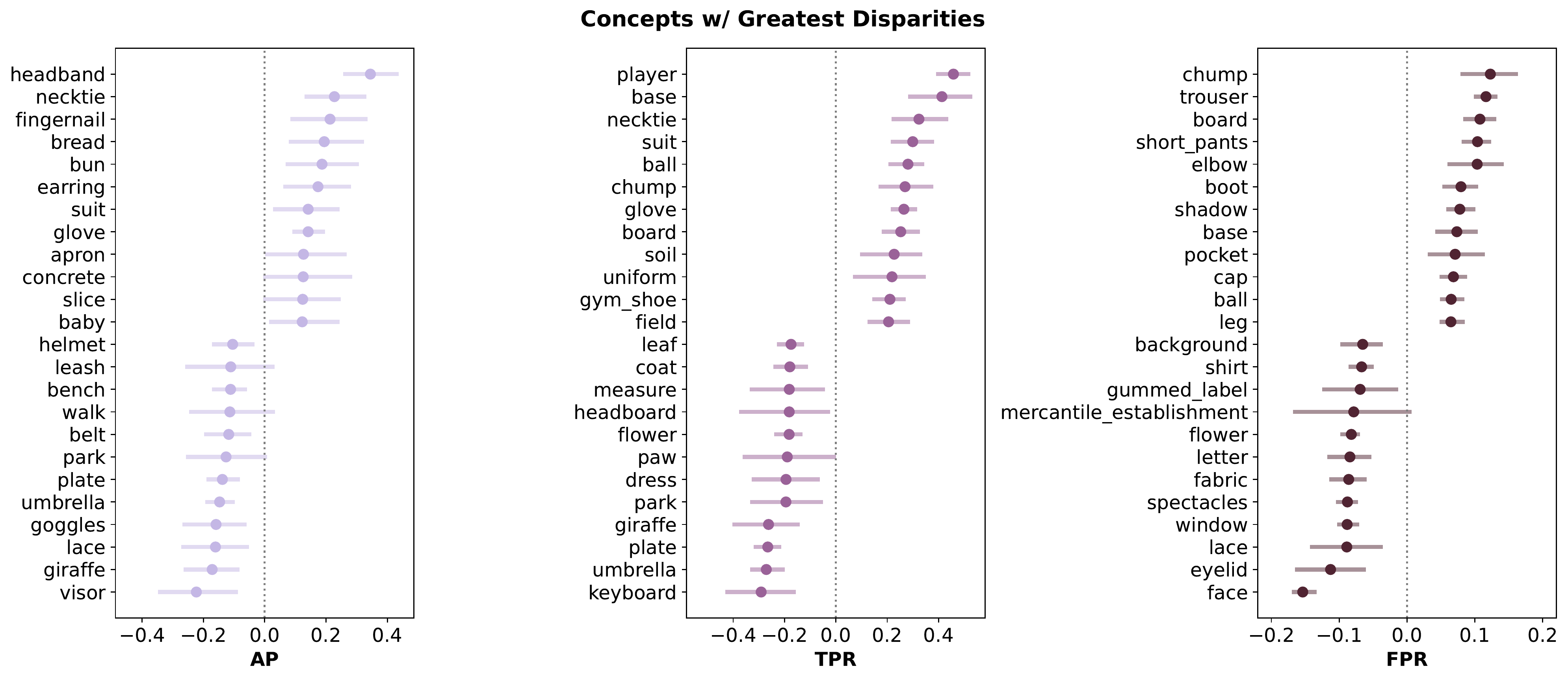}
     \end{subfigure}
     \hfill
     \caption{
     Disparities between \gender groups in Case Study \#1.
     \textbf{Left:} The \finaleval shows a significant difference in meanAP, which is obscured with the \texttt{Baseline Evaluation}, for both the Visual Genome and MS-COCO.
     \textbf{Right:} With the \finaleval there remain many concepts with large disparities. A positive disparity means that the model has a higher value for the \texttt{man} group than the \texttt{woman} group.
     Synset part of speech notation is removed to save space.
     The error bars are 95\% percentile intervals constructed with 250 bootstraps.
     }
\label{fig:case1_greatest_disp}
\end{figure}

We now discuss variations in disparities between \texttt{Evaluation Versions} across our metrics (\textit{Choice D}). Figure \ref{fig:case1_path_eval} shows concepts with the greatest changes in measurements for the Visual Genome.

First, we find that FPR and TPR are more susceptible than AP to the choice of terms and bounding box sizes used for operationalizing groups (\textit{Choice B}).
For example, disparities in TPR and FPR for concepts like \texttt{base.n.01} and \texttt{lace.n.01}, respectively, increase when filtering out small \gender bounding boxes, and disparities in FPR for \texttt{wristlet.n.01}
have significant and then insignificant demographic disparities as the filtering changes.
Upon further inspection, we find that variation in thresholds caused by the filtering of samples may explain these disparities (\textit{Choice E}).
Furthermore, disparities in some animal annotations like \texttt{bear.n.01} shrink when filtering out parent and child annotations that are often annotated for animals.

Second, TPR and FPR are less affected by correlations between prevalence and groups (\textit{Choice G}) than AP.
For AP, many concepts that initially showed better performance for group \texttt{woman} (such as \texttt{umbrella.n.01}, \texttt{dress.n.01}, and \texttt{hair.n.01}) have smaller or no disparities when using a fixed sampling ratio.
Disparities for some concepts like \texttt{earring.n.01} and \texttt{helmet.n.02} switch, showing better performance for the other group.
Furthermore, conclusions about disparities in average precision over \textit{all} concepts change between evaluation methods (left-side of Figure \ref{fig:case1_greatest_disp}): the \texttt{Baseline Evaluation} does not show a significant disparity between the two groups, whereas the \finaleval method shows better performance for group \texttt{man}.
Patterns are similar for MS-COCO, shown in Appendix \ref{app:cs1_res}.

In Section \ref{sec:root_cause}, we discuss large disparities in the \finaleval (right-side of Figure \ref{fig:case1_greatest_disp}).

This Case Study shows the important relationship between average precision and sampling: when measuring with average precision, the magnitude and direction of disparities for many concepts are greatly affected when controlling for the ratio of positive and negative samples between groups.
Furthermore, while the \texttt{Evaluation Baseline} shows no disparity in overall performance between groups, a notable disparity appears when controlling for the proportion of positive and negative samples between groups.
Finally, the terminology and bounding box size used for operationalizing \gender affects the existence and size of disparities for individual concepts.

These findings suggest that practitioners should aim for similar prevalences between groups when performing measurements with average precision.
In addition, when assigning groups based on existing annotations we recommend performing manual inspections to ensure labels were applied in a way that is representative of the real-life groups that are the focus of measurement and that annotators used the terms with consistency according to information present in the image.
Finally, disaggregating measurements across individual concepts is necessary for identifying large disparities that can be masked by summary metrics.

\section{Case Study \#2: Evaluating Multiple Models across Training Paradigms}
\label{sec:case2}
For this second case study, we perform a comparison of geographic disparities using a vision transformer and convolutional neural network.

\subsection{Evaluation set-up}

Following \cite{goyalfairness2022,goyal2022vision}, we select models that are both pre-trained and fine-tuned with ImageNet-22K but have different architectures.
For the vision transformer we use ViT \cite{https://doi.org/10.48550/arxiv.2010.11929} and for the convolutional model we use TResNet \cite{tresnet}, both optimized for ImageNet-22K classes following \cite{ridnik2021imagenet21k}.  We explore how evaluation choices using the Dollar Street dataset affect disparity findings both at the \textit{individual} model level and \emph{between} models.

\begin{figure}
\centering

    \begin{subfigure}{0.5\linewidth}
        \centering
        \raisebox{2mm}{
        \includegraphics[width=\textwidth]{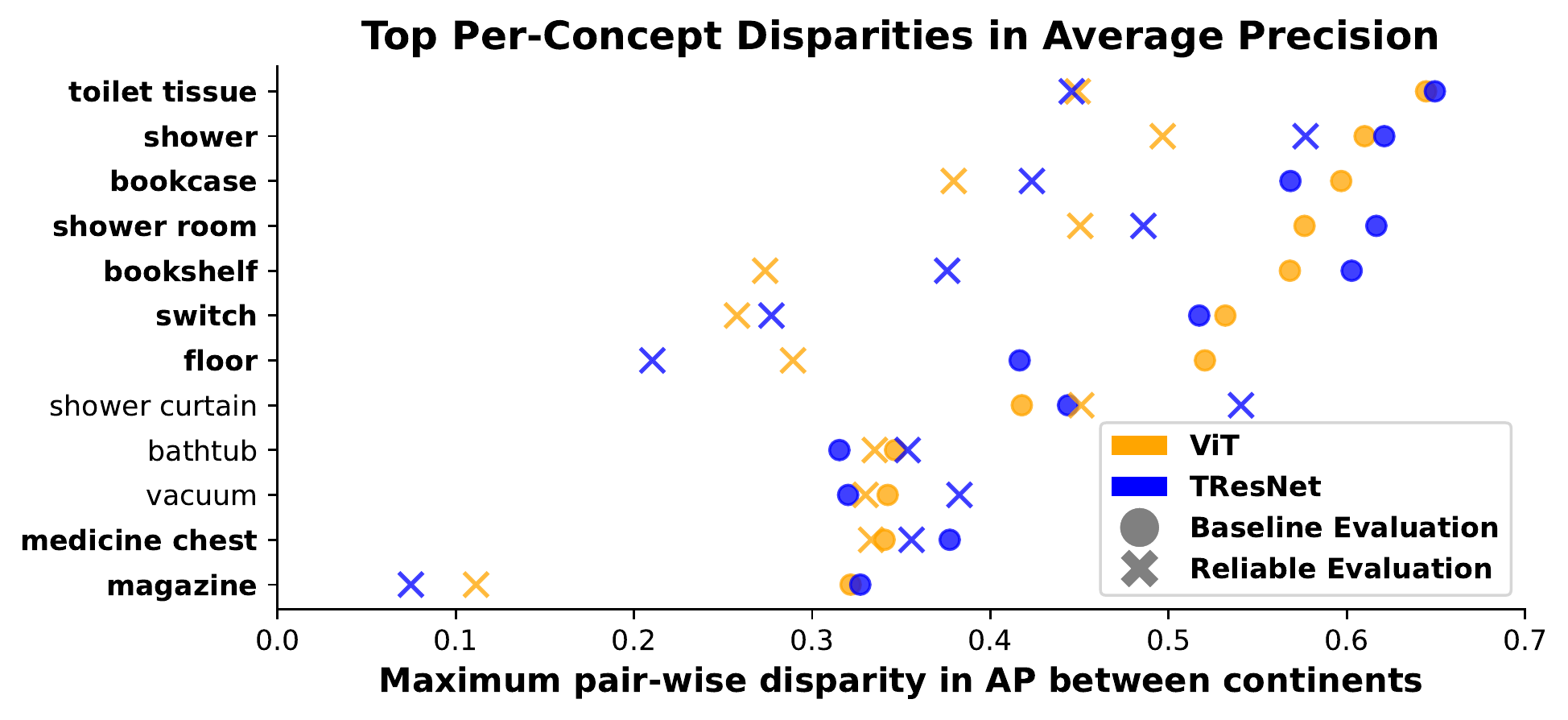}}
    \end{subfigure}  \rulesep
    \begin{subfigure}{0.35\linewidth}

        \centering
                          \includegraphics[width=\textwidth]{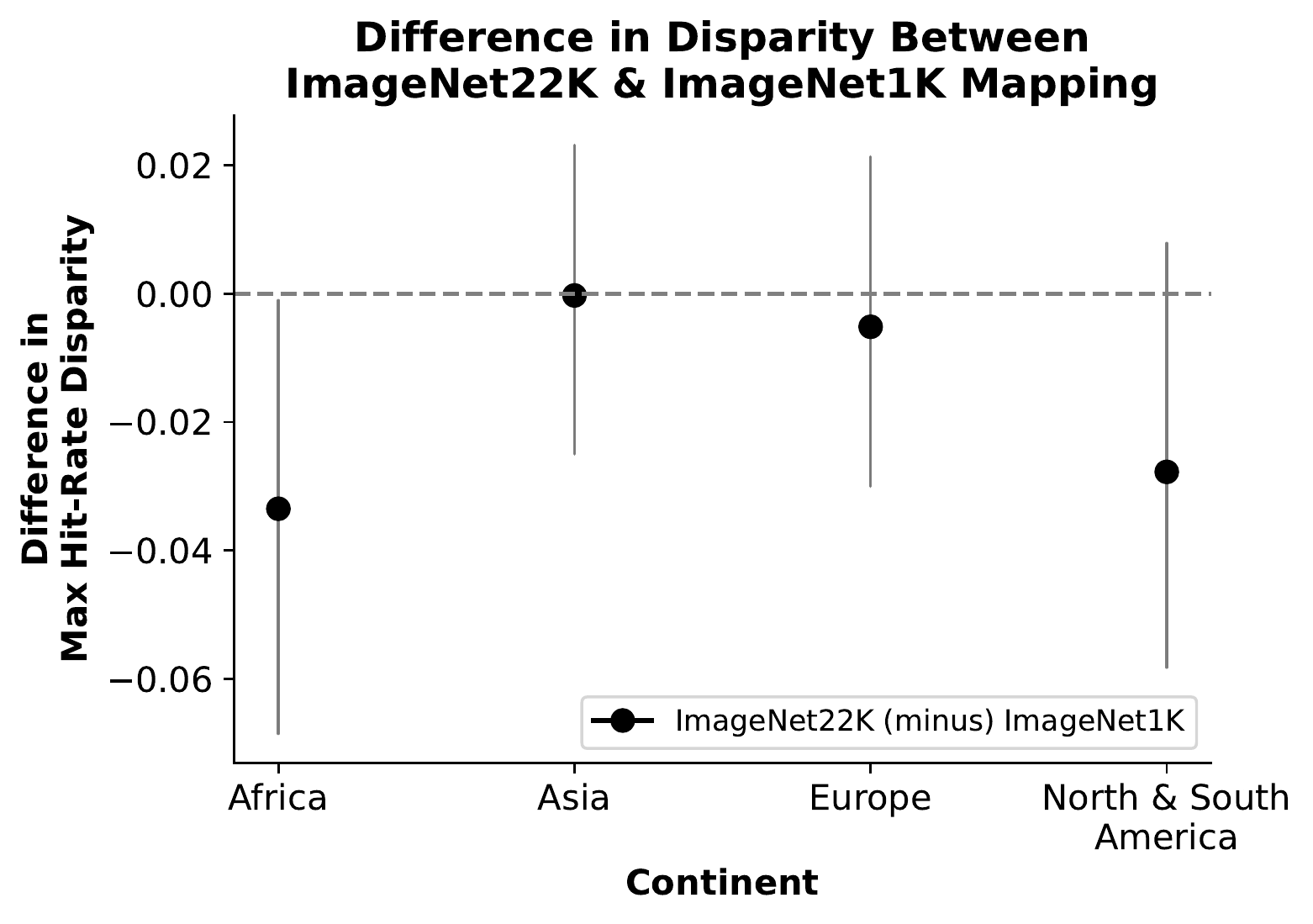}
    \end{subfigure}
\caption{
Disparities between continents in Case Study \#2.
\textbf{Left:} The per-concept disparities between continents decrease for many concepts (shown in \textbf{bold}) across both models when using a fixed ratio of positive and negative samples between groups, as shown by the leftward shift of disparities between the \texttt{Baseline Evaluation} and the \finaleval.
This also affects the extent to which one model is better than the other for some concepts, shown by the switch in ordering between blue and orange markers for concepts like \texttt{bookcase}.
\textbf{Right:} Disparities for ViT decrease for some continents when mapping Dollar Street to ImageNet-22K classes rather than ImageNet-1K classes.
The error bars are 95\% percentile intervals constructed with 250 bootstraps.
}
\label{fig:case2}
\end{figure}

\vspace{1mm}

\noindent \textbf{Data selection}

\vspace{1mm}

\noindent \textit{\textbf{Choice A: Data source.}}
We use the Dollar Street \cite{dollarstreet} dataset, which has been used extensively to evaluate models for biases across geographic regions \cite{devries2019does, Singh_2022_CVPR, goyalfairness2022, rojas2022dollar}.
It contains images of people and objects across the globe that are labeled based on the country where they were taken.
Some images are duplicated with multiple labels; we collapse these so that each image only appears once, creating a multi-label classification set-up.
We remove all images without labels.

\vspace{1mm}

\noindent \textit{\textbf{Choice B: Operationalization of groups.}}
Following \cite{goyalfairness2022}, we focus on the continent where the photo was taken and map countries to continents and create groups: Africa, Asia, Europe, and the Americas (North and South America).
Because the country information is provided by the photo uploader, we have high confidence that the group information is correct and therefore do not perform any filtering based on group information.

\vspace{1mm}

\noindent \textit{\textbf{Choice C: Data \& model class mappings.}}
Because these models are not zero shot, we need to map the Dollar Street labels to the object classes the models were trained on.
We perform two baseline evaluations with different mappings from Dollar Street to ImageNet: \texttt{Baseline Evaluation (ImageNet-22K)}, where we map to ImageNet-22K classes following \cite{goyalfairness2022}, and \texttt{Baseline Evaluation (ImageNet-1K)}, where we map to ImageNet-1K classes following \cite{rojas2022dollar}.
Examples of the mappings for both ImageNet-[22/1]K classes are shown in Appendix \ref{app:cs2_mappings}.
These models predict only a subset of the ImageNet classes \citep{ridnik2021imagenet21k}, so we remove any incompatible concepts.

\vspace{1mm}

\noindent \textbf{Performance metric}

\vspace{1mm}

\noindent \textit{\textbf{Choice D: Choosing the family of metrics.}}
We evaluate disparities using hit-rate, where an image is a \emph{hit} if one of the top-5 predictions is correct \citep{goyalfairness2022}.
Because hit-rate does not extend to per-concept measurements and the number of concepts per image varies, we also use average precision (AP).

\vspace{1mm}

\noindent \textit{\textbf{Choice E: Ranking vs classification metrics.}} Both hit-rate and AP are ranking metrics and do not require a threshold.

\vspace{1mm}

\noindent \textbf{Data sampling}

\vspace{1mm}

\noindent \textit{\textbf{Choice F: Filtering rare labels.}}
Dollar Street's small size and large quantity of groups introduces a tradeoff between coverage and noise.
Initially, we follow \cite{goyalfairness2022} and evaluate all concepts.
However, to reduce noise in our measurements, we remove concepts with less than 30 samples per group.

\vspace{1mm}

\noindent \textit{\textbf{Choice G: Controlling for correlations.}}
We initially follow \cite{goyalfairness2022,rojas2022dollar,devries2019does} and do not perform additional sampling.
However, in our \finaleval we follow the sampling method described in Case Study \#1 with a sampling ratio of 1:4.
We construct confidence intervals using bootstrapping, as described in Appendix \ref{app:cs2_ci}.

\subsection{Results}

We investigate how different \texttt{Evaluation Versions} affect disparity findings.

The left-side of Figure \ref{fig:case2} shows that the comparison of disparities between ViT and TResNet are affected by correlations between prevalence and groups (\textit{Choice G}) and that disparities tend to decrease across both models with the \finaleval method.
Furthermore, the degree to which one model is better than the other can change between evaluation methods, as shown for concepts like \texttt{shower curtain} and \texttt{comb}.
For other concepts like \texttt{bookcase}, the model that has the greater disparity changes based on which evaluation method is used.
In Appendix \ref{app:cs2_rare}, we also show that without removing rare concepts (\textit{Choice F}), many of the largest disparities are for concepts with few samples.
Finally, the mappings between dataset labels and model classes (\textit{Choice C}) affect a single model's disparity findings.
The right-side of Figure \ref{fig:case2} shows the difference in per-continent disparity between the two Dollar Street to ImageNet mappings when evaluating ViT.
The magnitude of the performance disparities in ViT is larger for Africa when using the ImageNet1K class mappings than when using the ImageNet22K class mappings, indicating that the alignment of dataset classes to model classes can affect the magnitude of observed disparities.

This Case Study reinforces the importance of sampling when using average precision and shows that sampling also affects \emph{how much} one model is better than another.
Furthermore, the mapping of dataset labels to model classes affects the magnitude of observed disparities.

These findings suggest that practitioners should filter out concepts that have too few samples for each group when using Dollar Street.
In addition, multiple mappings between dataset and model classes and their tradeoffs between coverage and alignment should be considered carefully.

\section{Root Cause Analyses of Large-Disparity Concepts}
\label{sec:root_cause}

\NumTabs{40}
\begin{figure}[!htb]

         \begin{tabular}{p{15cm}}
        \small{\textbf{Low imageability:} Concept is not concrete and is hard to portray with visual representations} \\
        \small{\texttt{background.n.02} (Visual Genome)}
        \tab \tab \tab \tab \tab \tab \tab \tab \small{\texttt{measure.n.01} (Visual Genome)}\\
        \includegraphics[height=17mm]{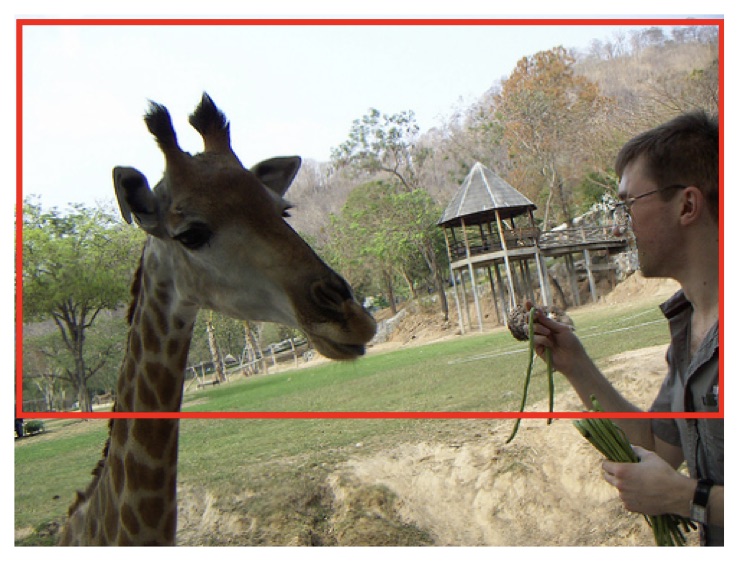}
        \includegraphics[height=17mm]{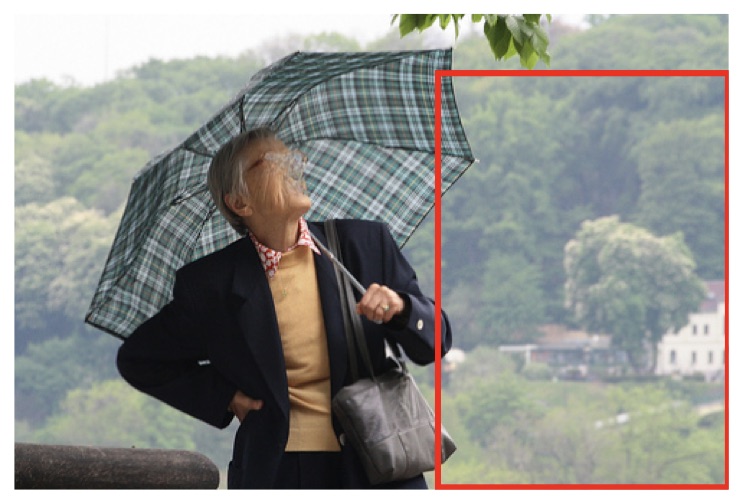}
        \includegraphics[height=17mm]{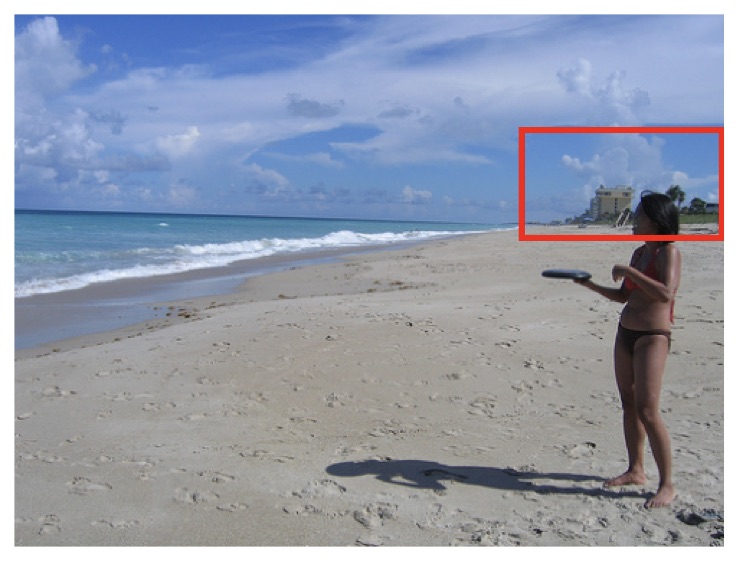}
                                \tab
        \includegraphics[height=17mm]{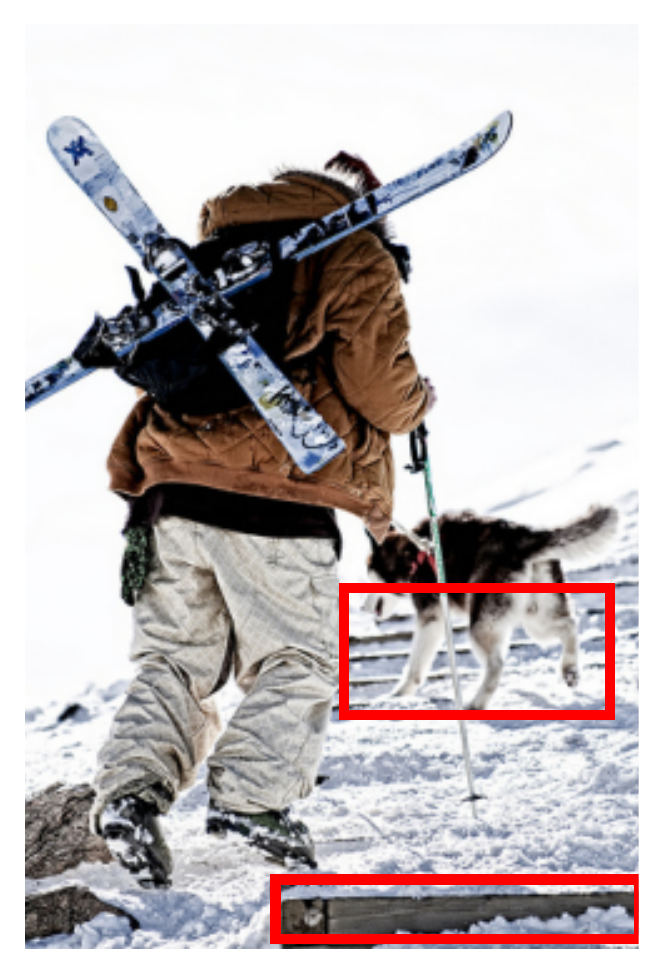}
        \includegraphics[height=17mm]{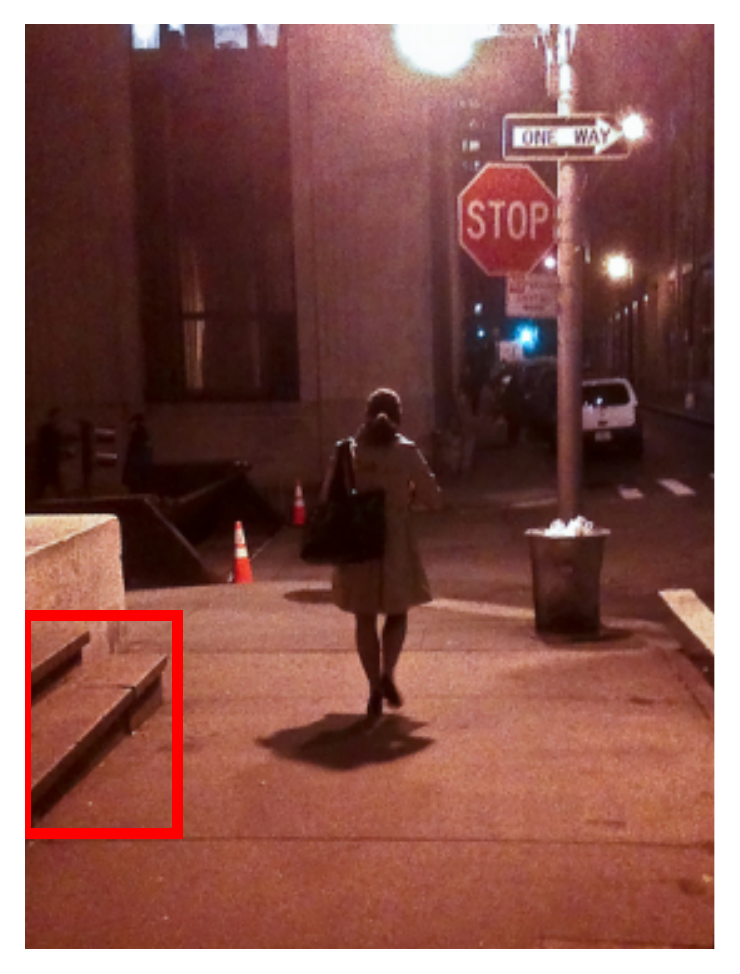}
         \includegraphics[height=17mm]{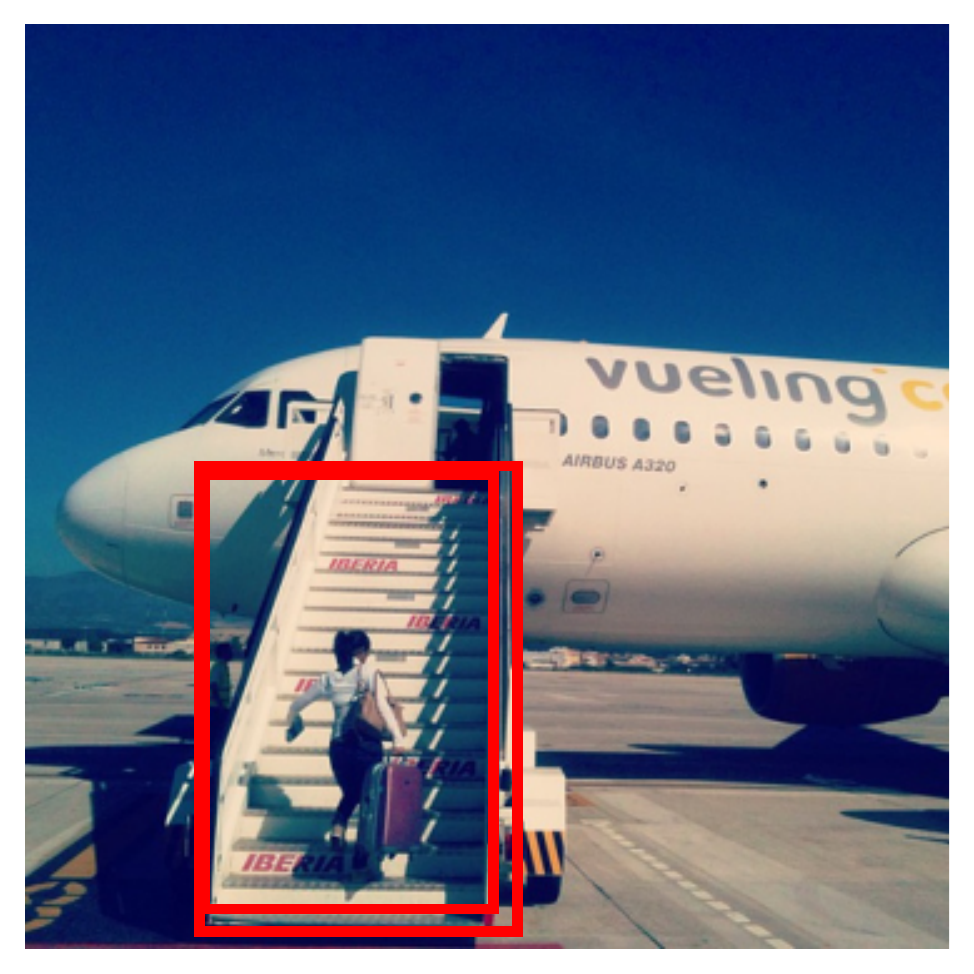}
  \end{tabular}

    \vspace{1mm}

  \begin{tabular}{p{15cm}}
        \small{\textbf{Indiscernible:} Cannot determine if concept is present and may encourage use of biased shortcuts}\\
        \small{\texttt{dress.n.01} (Visual Genome)}
        \tab {\small{\texttt{floor} (Dollar Street)}}
        \tab \tab \tab {\small{\texttt{handbag} (COCO)}}\\
        \includegraphics[height=17mm]{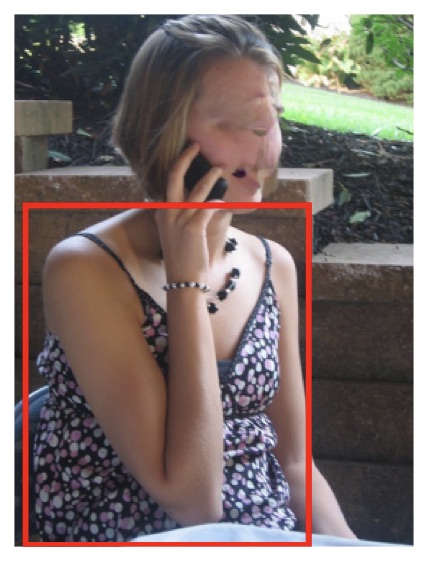}
        \includegraphics[height=17mm]{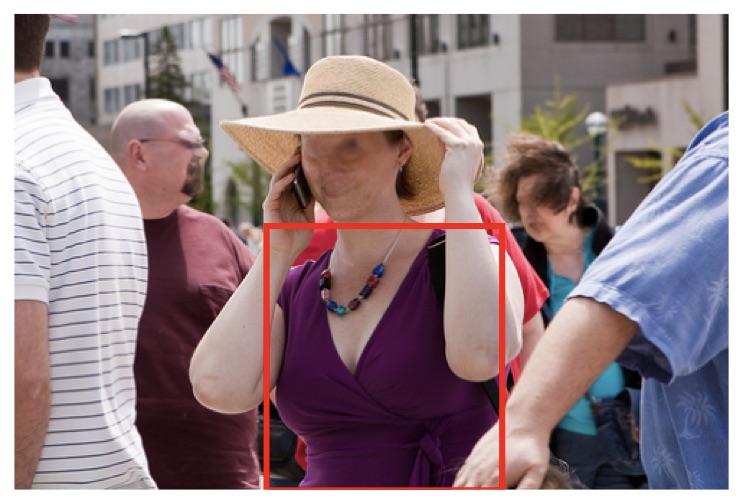}
        \tab
        \includegraphics[height=17mm]{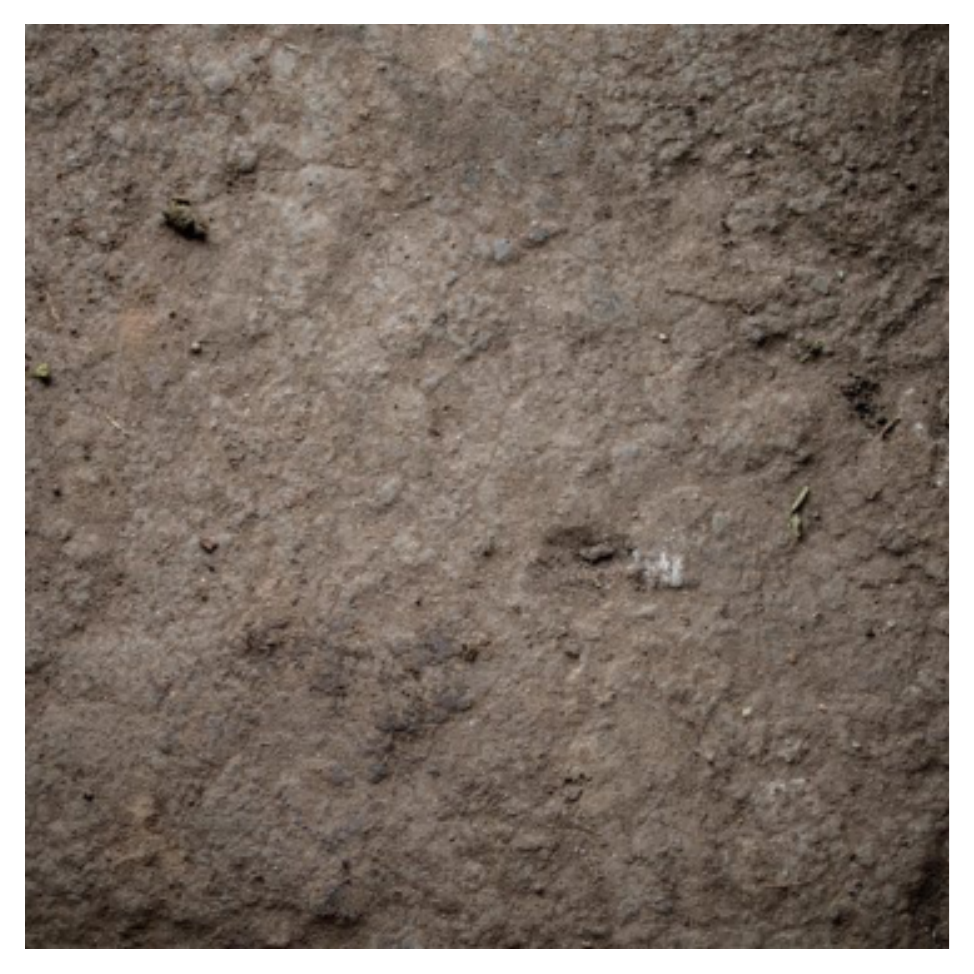}
        \includegraphics[height=17mm]{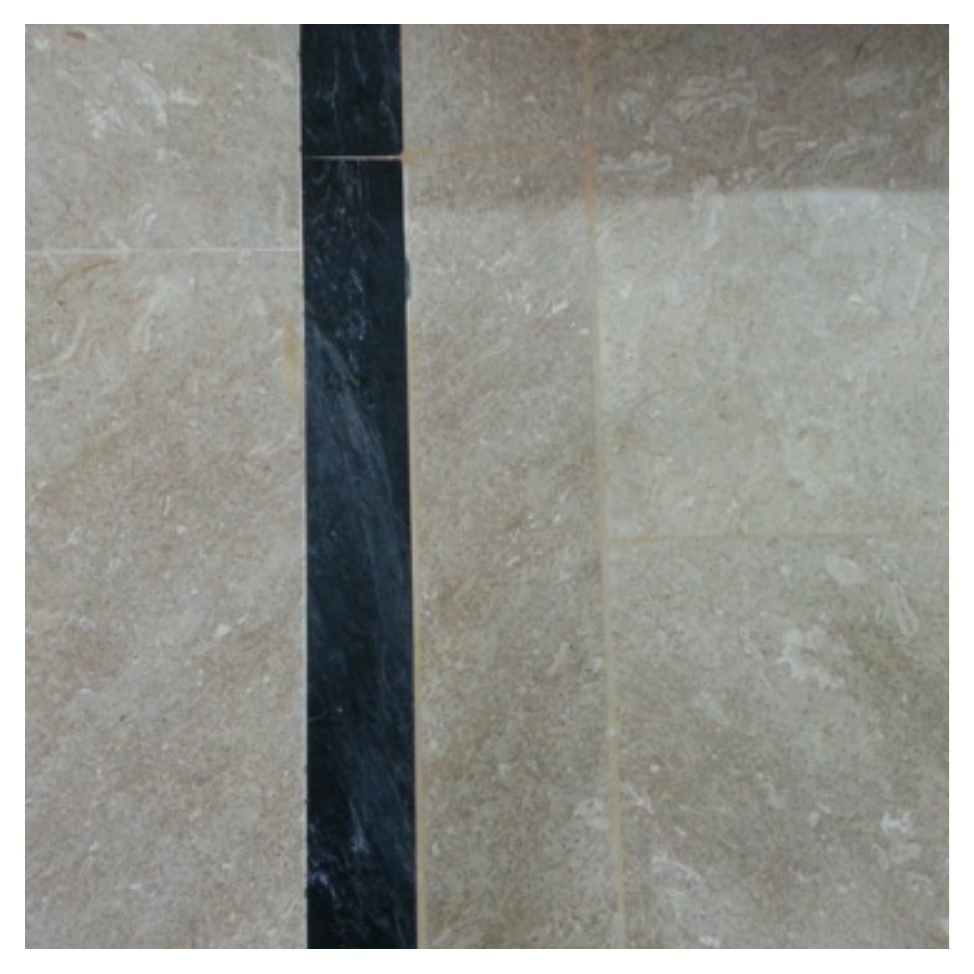}
        \tab
        \includegraphics[height=17mm]{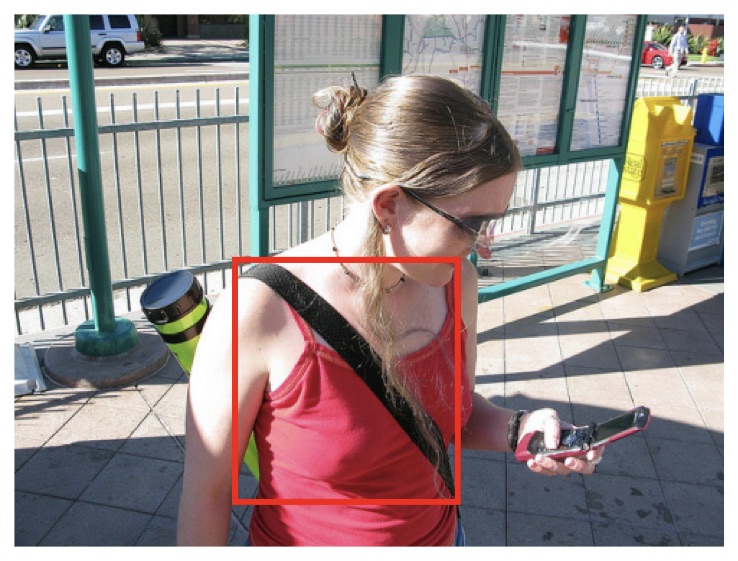}
        \includegraphics[height=17mm]{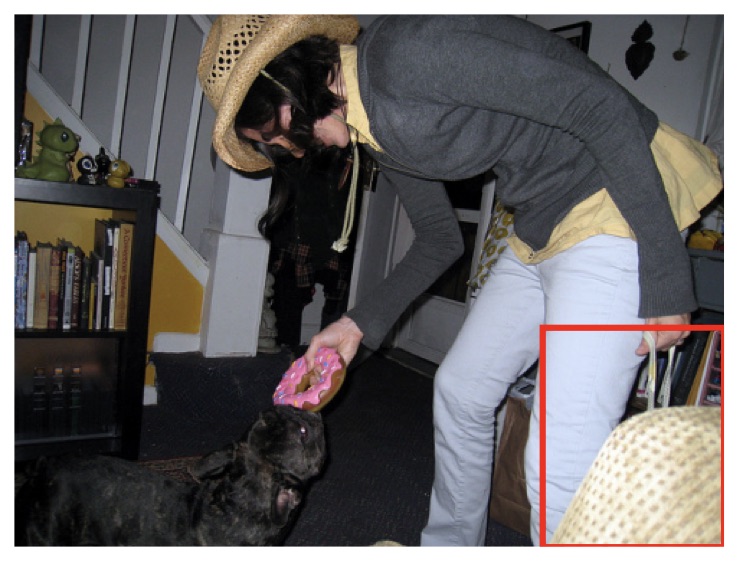}
  \end{tabular}

  \vspace{1mm}

    \begin{tabular}{p{15cm}}
        \small{\textbf{Ambiguous definitions:} Labels capture multiple valid definitions and can vary between groups }\\
        \small{\texttt{base.n.01} (Visual Genome)}
        \tab \tab \tab \tab \small{\texttt{shower} (Dollar Street)}
        \tab \tab \small{  \texttt{tie} (COCO)}
        \\
        \includegraphics[height=17mm]{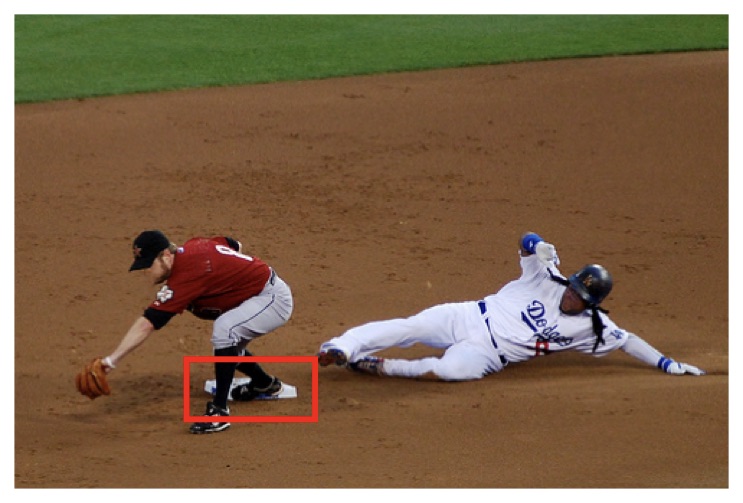}
        \includegraphics[height=17mm]{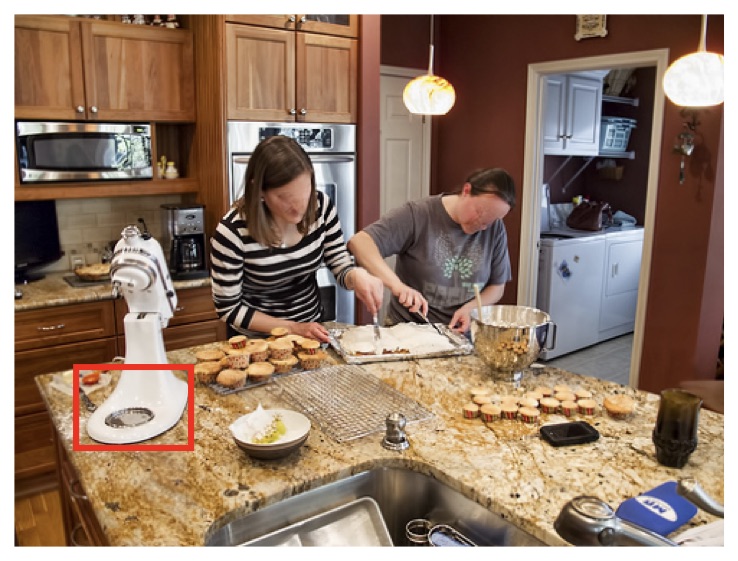}
        \includegraphics[height=17mm]{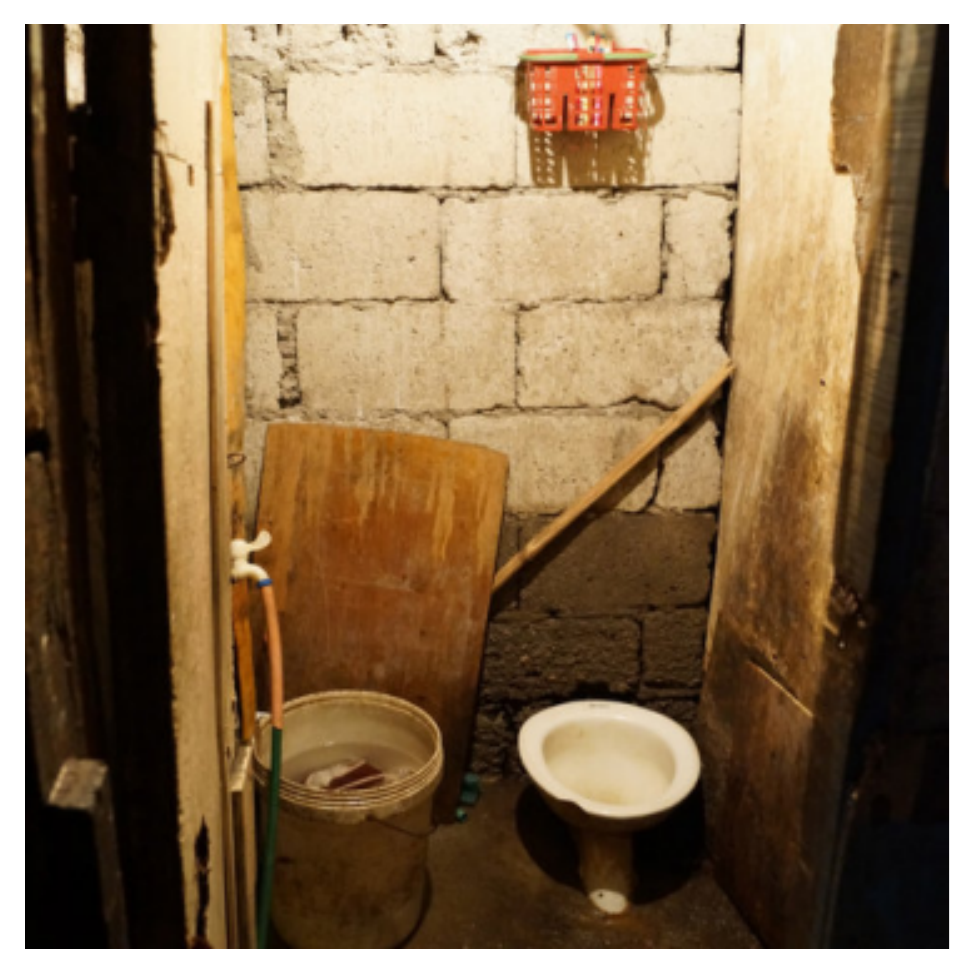}
        \includegraphics[height=17mm]{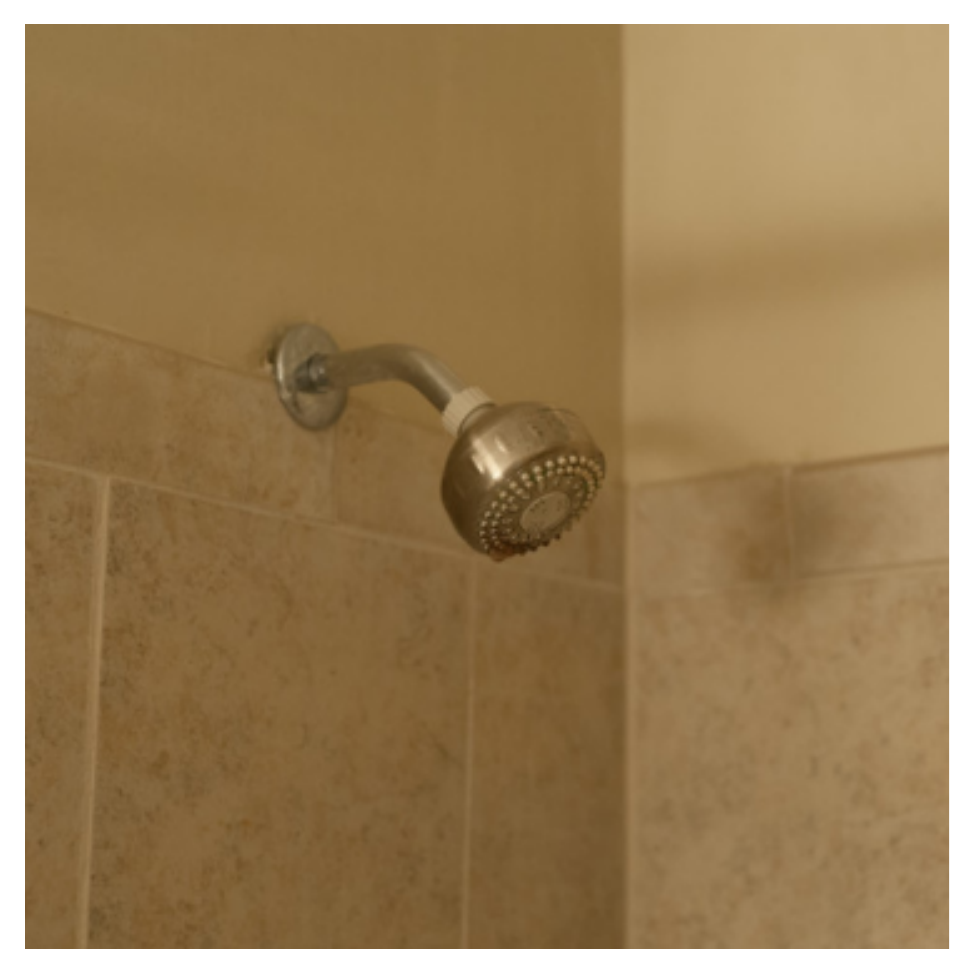}
        \includegraphics[height=17mm]{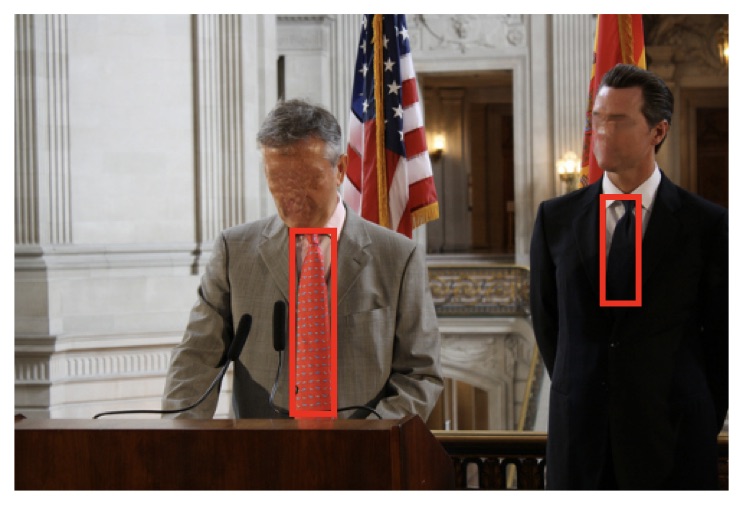}
        \includegraphics[height=17mm]{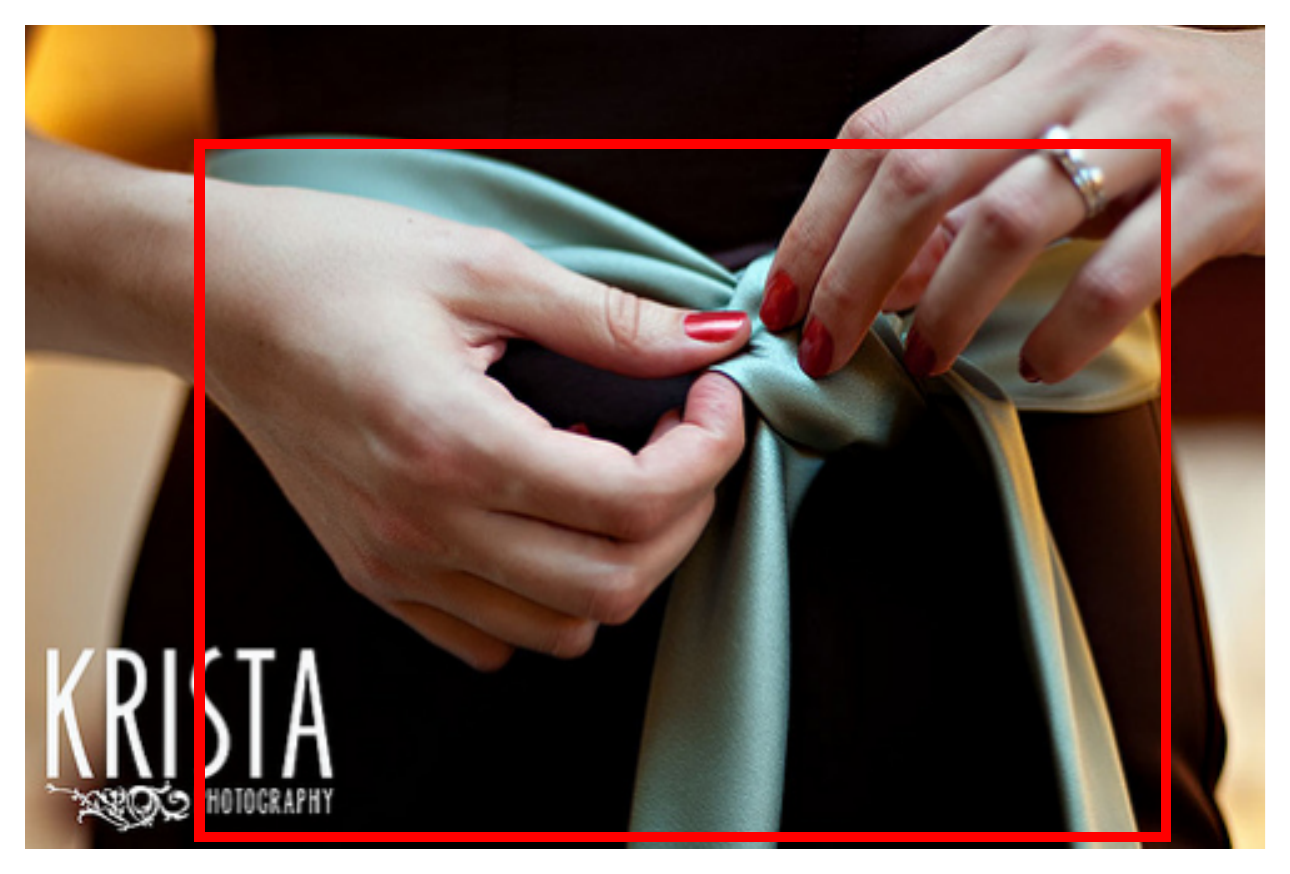}
  \end{tabular}

  \vspace{1mm}

    \begin{tabular}{p{15cm}}
        \small{\textbf{Incorrect labels:} Pictured objects do not align with concept definition, which may reflect the majority} \\
        \small{\texttt{necktie.n.01} (Visual Genome)}
        \tab  \small{  \texttt{toilet\_paper} (Dollar Street)}
        \tab \small{\texttt{spectacles.n.01} (Visual Genome)}\\
        \includegraphics[height=17mm]{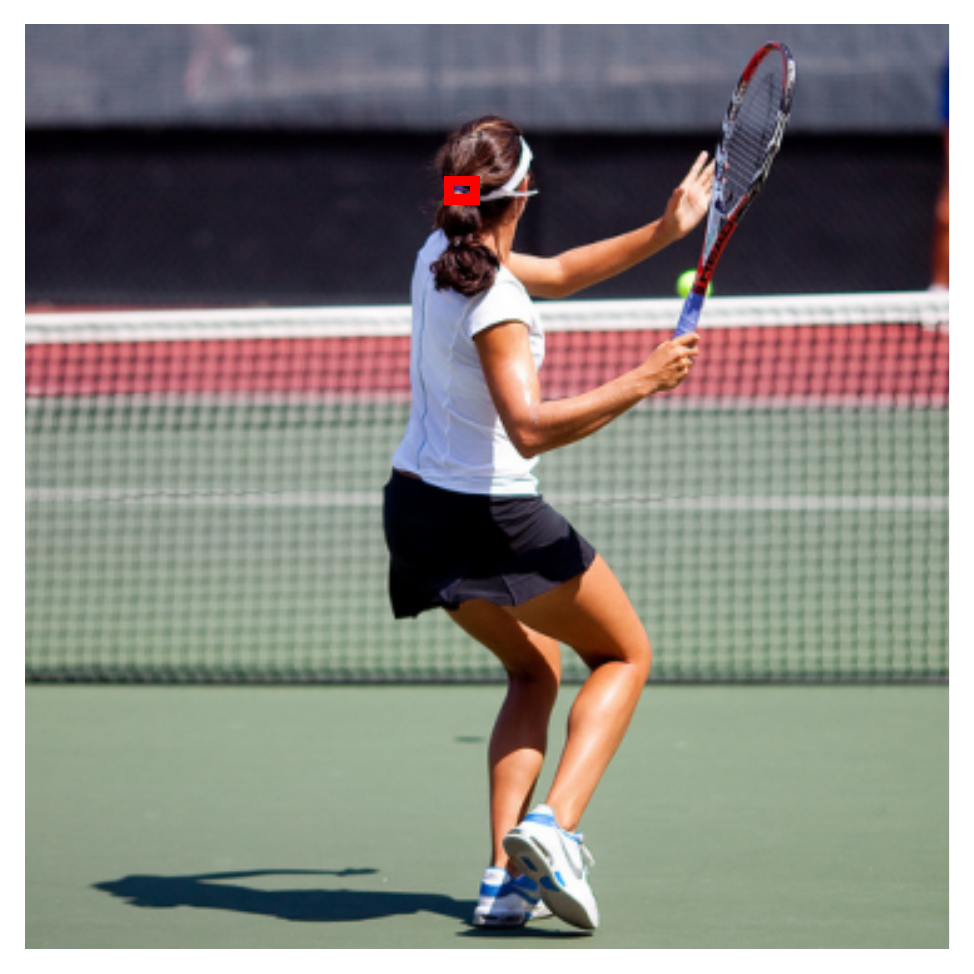}
        \includegraphics[height=17mm]{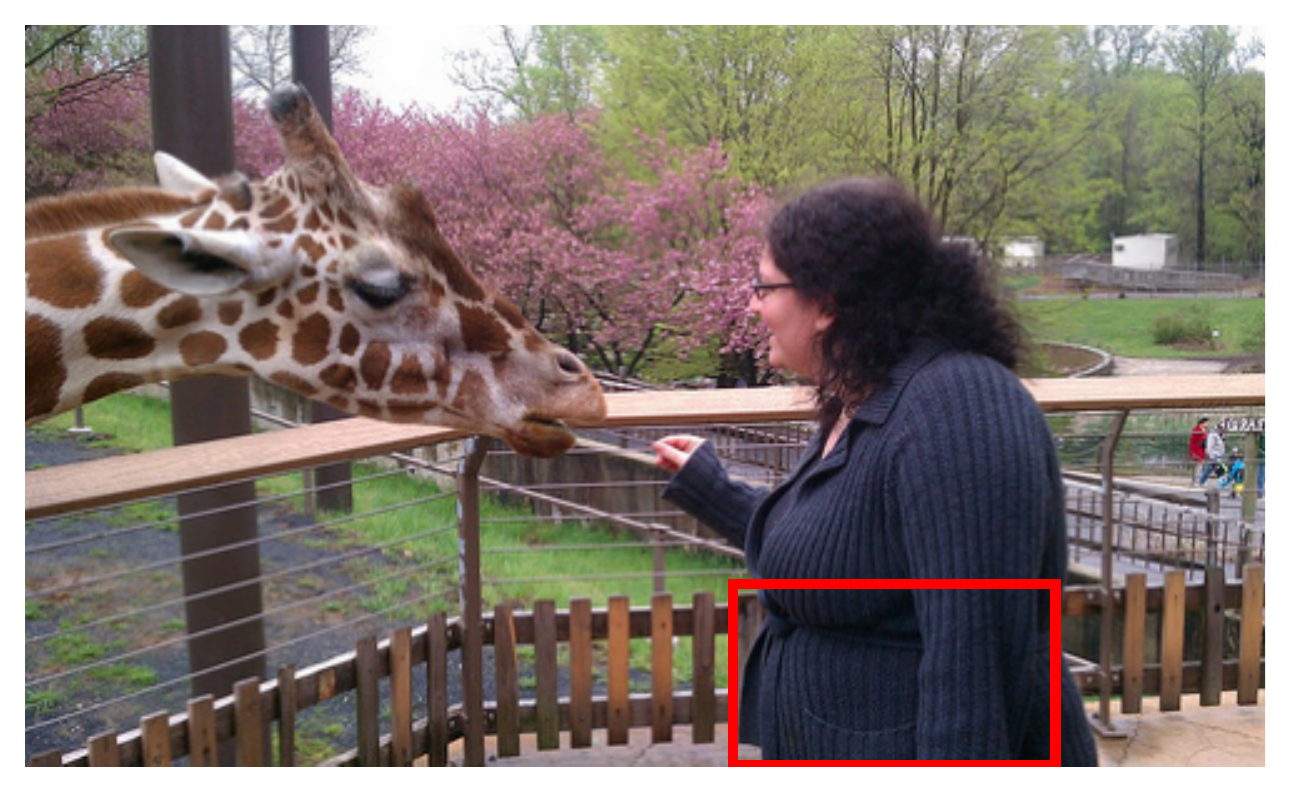}
        \includegraphics[height=17mm]{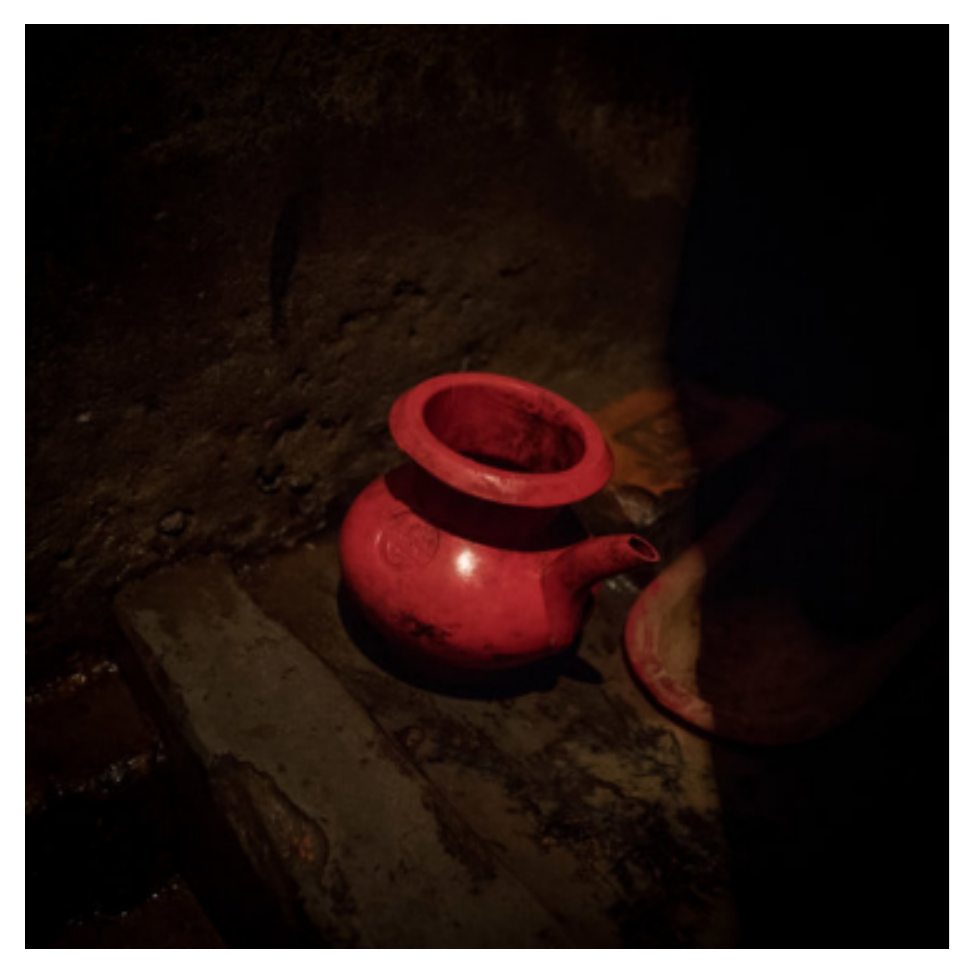}
        \includegraphics[height=17mm]{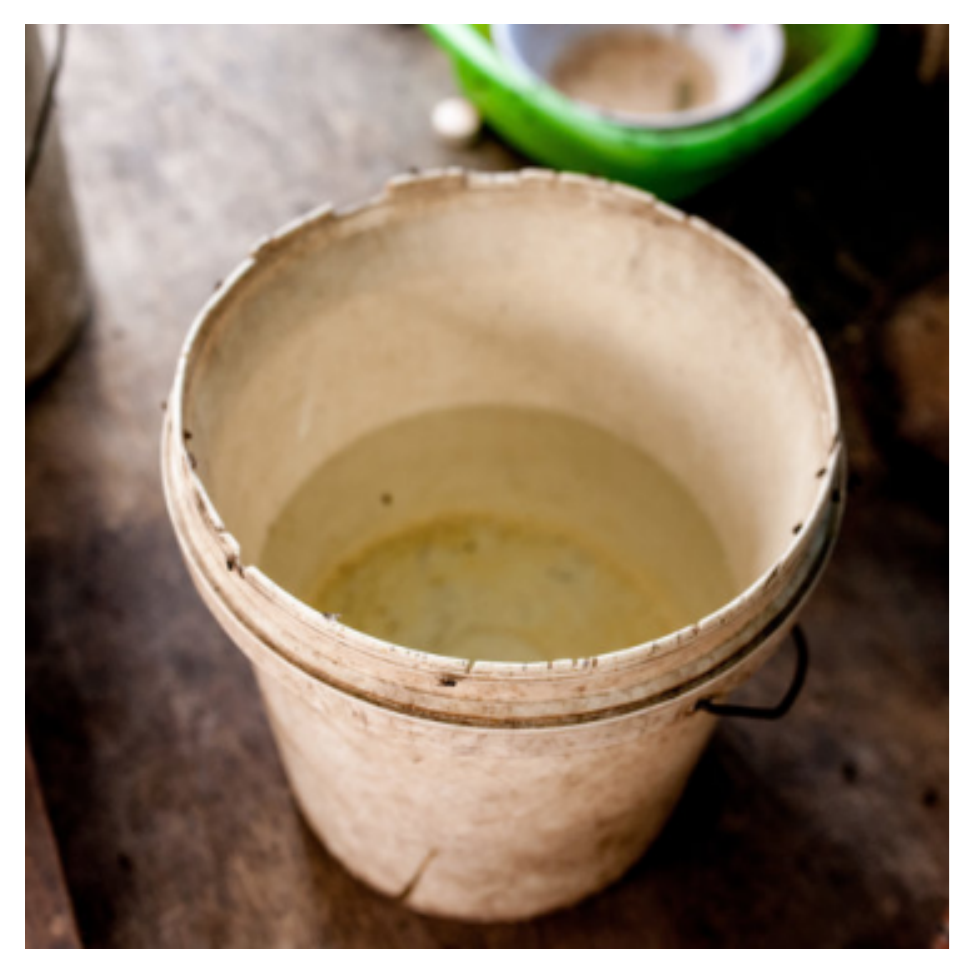}
        \tab \tab
        \includegraphics[height=17mm]{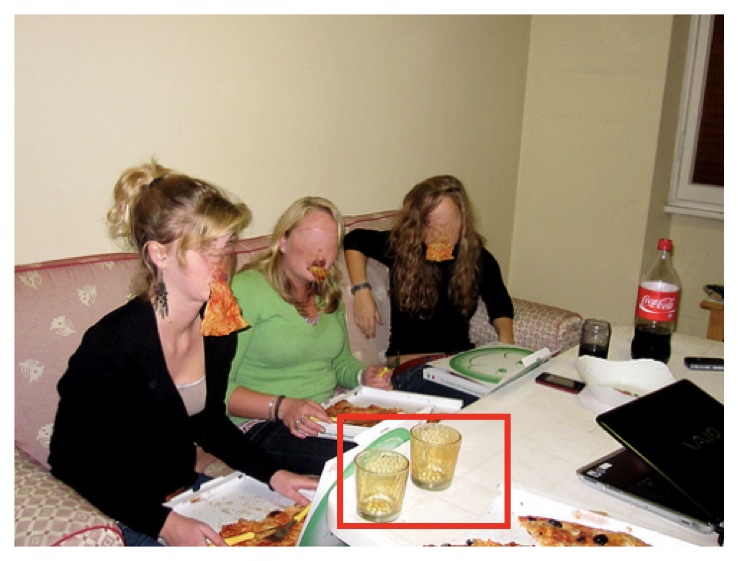}
        \includegraphics[height=17mm]{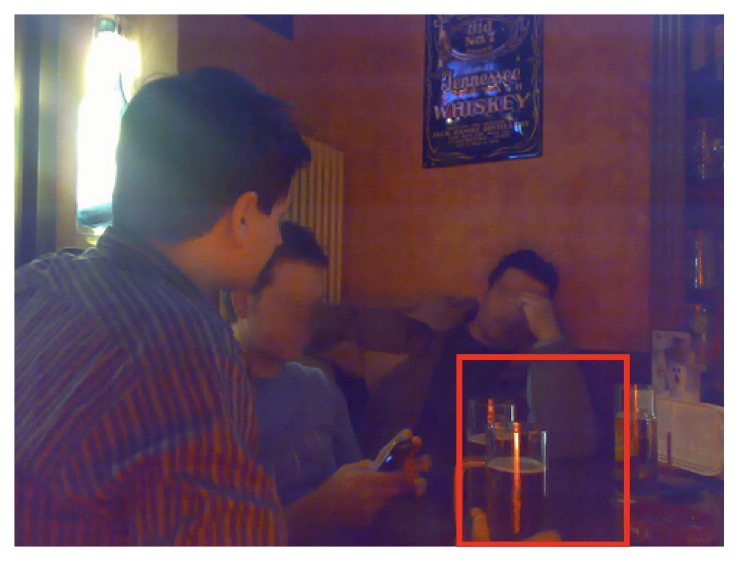}
  \end{tabular}

\caption{Concepts with large disparities between groups contain complexities in representation that affect groups differently. These patterns occur across datasets. We show in \textcolor{red}{RED} bounding box annotations and blur faces for the privacy of those depicted. We use the original images in our evaluations.}
\label{tbl:model_mistakes_labels}

\end{figure}

In the previous two sections, we showed how choices employed in evaluations affect observed demographic disparities in CLIP, ViT, and TResNet.
These investigations led to more \finaleval methods, which themselves have trade-offs, as discussed in Section \ref{sec:choices}.
However, they can provide high-level insights about model performance between groups.
In this Section, we qualitatively investigate concepts with large disparities in performance observed with our \finaleval measurements (shown in Figures \ref{fig:case1_greatest_disp}, \ref{fig:case2}, \&  \ref{fig:case1_greatest_disp_coco}).
We find that they can have multiple or malleable definitions, especially between groups.
Often, representations have \textbf{low imageability}, are \textbf{indiscernible}, have \textbf{ambiguous definitions}, or contain \textbf{incorrect labels}.
These patterns are not mutually exclusive and can affect demographic groups differently.
Figure \ref{tbl:model_mistakes_labels} summarizes these variations, and we discuss more below.

First, some concepts have \textbf{low imageability}, \ie they are difficult to portray with visual representations.
This relates to ``concreteness,'' which measures the amount that a concept refers to a perceptible entity.
Using an existing ranking of concreteness among 40,000 English lemmas \cite{brysbaert2014concreteness}, we observe that some concepts with large performance disparities have lower concreteness: \texttt{background.n.02} and \texttt{measure.n.01} have concreteness ratings of 3.26/5.00 and 3.59/5.00, respectively, while most objects have a rating above 4.50/5.00.
Figure \ref{tbl:model_mistakes_labels} shows that backgrounds are not consistently delineated, and photos for \texttt{measure.n.01} often represent stair steps rather than a ``maneuver made as part of progress toward a goal'' \citep{miller-1994-wordnet}.

Second, whether the concept is present in the image may be \textbf{indiscernible}, forcing the viewer to use other context in the image.
This could lead to bias amplification \cite{zhao2017men, DBLP:journals/corr/abs-2201-11706,https://doi.org/10.48550/arxiv.2102.12594}.
For example, annotators label concepts \texttt{dress.n.01} and \texttt{handbag} in images containing \texttt{woman} annotations, even when it is impossible to discern between a dress and a shirt, or a handbag and straps, respectively.
Even if the image source provides labels, the image may not capture whether the concept is present: many photos of \texttt{floors} in Dollar Street do not contain grounding visuals like walls or furniture.

Third, some concepts have \textbf{ambiguous definitions} and labelers employ multiple distinct, valid interpretations of concrete concepts.
This often happens along group lines, where the same label is used for different objects for different groups.
For example, annotations for the concept \texttt{base.n.01} tend to more frequently capture a baseball field base for images within the \texttt{man} group and the structural foundation of an object for images with the \texttt{woman} group. \texttt{Tie} includes decorative ties around the waist when co-occurring with the \texttt{woman} group but rarely with the \texttt{man} group.
Within Dollar Street, approximately 30\% of images of \texttt{showers} taken in Asia show a bucket or basin rather than an elevated shower head or hose, as opposed to less than 5\% of images taken in Europe.

Finally, some concepts have \textbf{incorrect labels} where the pictured object does not align with the concept definitions.
The labels may misalign more frequently for minority groups or represent the perspective of an advantaged group.
For example, annotations of \texttt{necktie.n.01} include hair-ties when co-occurring with the \texttt{woman} group much more than with the \texttt{man} group, and photos from Asia of \texttt{toilet\_paper} often feature water sources devoid of paper sheets.
If one considers hair-ties distinct from neckties, and toilet paper different than buckets of water, then in these cases a model's observed performance disparity may relate more to a misalignment between a concept definition and a group than its inability to identify variants of the same object between two groups.

Furthermore, Appendix \ref{app:qual} shows that the Visual Genome and MS-COCO often depict the \texttt{woman} group in art or historical pieces, which can affect the classification difficulty between groups \cite{DBLP:journals/corr/abs-1802-03601}.

Our analyses suggest possible mitigations to improve performance between groups.
For example, in training datasets, developers can clarify guidelines regarding missing information in images so that labelers need not rely on (potentially biased) inferences based on incomplete visual references.
In addition, label definitions can be made more explicit to avoid ambiguous or incompatible meanings, or expanded to account for groups beyond predominant perspectives. Classes with low imageability may also be removed from model training if found to be unreliably annotated.
When used for evaluation datasets, these changes could also improve the reliability of measurements.

\section{Conclusion}
\label{sec:discussion}

In this work, we outlined key choices and tradeoffs for disaggregated evaluations of multi-label object recognition models across demographic groups and showed that they have fundamental complexities not seen in single concept binary classifiers.

We first described important design choices that span data selection, performance metrics, and sampling.
These choices require thoughtful consideration of many tradeoffs, including annotation cost and quality, group coverage and specificity, and confounders between groups and label distributions.
In two Case Studies, we showed that these evaluation choices affect the existence, magnitude, and direction of observed disparities for individual concepts and in aggregate.
This yielded actionable guidance for practitioners, such as fixing the ratio of positive and negative samples between groups when using average precision and performing mappings between dataset and model labels with care.
Finally, we identified potential root causes of disparities that exist even with more reliable evaluations, including low imageability concepts and ambiguous definitions, and suggested mitigations such as improved labeling guidelines for missing visual references.

This work captures only a subset of the decisions that can affect disparity analyses of multi-label, multi-class computer vision models.
There remain many challenges in performing these evaluations, including the aforementioned complexities in concept definition and representation as well as non-trivial trade-offs involved in operationalizing groups and controlling for correlations between co-occurring labels and groups.
As practitioners work towards identifying and reducing disparities in performance of multi-label object recognition models and image classifiers, we hope this is a meaningful step towards more reliable and insightful evaluations.

\section{Acknowledgments}

We thank Aaron Adcock, Alex Vaughan, Ishan Misra, Enpeng Yuan, Hailey Nguyen, Tsung-Yu Lin, Mannat Singh, Levent Sagun, Hervé Jegou, and Megan Richards for their feedback on this work.

\newpage

\bibliographystyle{ACM-Reference-Format}
\bibliography{references}
\newpage
\section{Appendix}
\label{sec:appendix}

\subsection{Additional Details: Case Study \#1}

\subsubsection{Gender-based terms.}
\label{app:cs1_terms}

The synsets and caption terms used for adapting the Visual Genome and MS-COCO datasets to operationalize \gender groups are shown in Table \ref{tab:gendered}.
These terms are selected from previous work \cite{https://doi.org/10.48550/arxiv.2301.11100}, so that we can investigate the effects of term-based assignment of group information on disparity evaluations.
\textbf{Bolded} terms are those that we exclude in \texttt{Evaluation V3}.

\begin{table*}[h!]
\centering
\begin{subfigure}{0.45\linewidth}
    \centering \textbf{Visual Genome}\\[2ex]
    \begin{tabular}{p{0.4\linewidth} | p{0.4\linewidth}}

     \parbox{\linewidth}{\centering terms for group\\\texttt{man}\\[0.8ex]}
     & \parbox{\linewidth}{\centering terms for group \texttt{woman}\\[0.8ex]} \\ [0.5ex]

     \hline
      man.n.01,\newline male\_child.n.01,\newline guy.n.01,\newline male.n.01,\newline groom.n.01,\newline husband.n.01,\newline grandfather.n.01,\newline \textbf{father.n.01},\newline \textbf{son.n.01},\newline boyfriend.n.01,\newline brother.n.01,\newline grandson.n.01,\newline groomsman.n.01,\newline ex-husband.n.01,\newline uncle.n.01,\newline godfather.n.01
      &
      maid.n.02,\newline woman.n.01,\newline girl.n.01,\newline lady.n.01,\newline female.n.01,\newline \textbf{mother.n.01},\newline lass.n.01,\newline \textbf{ma.n.01},\newline widow.n.01,\newline bride.n.01,\newline \textbf{daughter.n.01},\newline grandma.n.01,\newline granddaughter.n.01,\newline bridesmaid.n.01,\newline girlfriend.n.01,\newline sister.n.01,\newline wife.n.01,\newline female\_child.n.01,\newline white\_woman.n.01,\newline dame.n.01,\newline matriarch.n.01,\newline \textbf{mother\_figure.n.01},\newline dame.n.02,\newline great-aunt.n.01,\newline donna.n.01
      \\
     \end{tabular}
     \end{subfigure}
     ~
     \begin{subfigure}{0.45\linewidth}
         \centering
        \textbf{MS-COCO}\\[2ex]
        \begin{tabular}{p{0.4\linewidth} | p{0.4\linewidth}}

         \parbox{\linewidth}{\centering terms for group\\\texttt{man}\\[0.8ex]}
         & \parbox{\linewidth}{\centering terms for group \texttt{woman}\\[0.8ex]} \\

         \hline
         man,\newline mans,\newline men,\newline boy,\newline boys,\newline \textbf{father},\newline \textbf{fathers},\newline \textbf{son},\newline \textbf{sons},\newline he,\newline his,\newline him

         & woman,\newline womans,\newline women,\newline girl,\newline girls,\newline lady,\newline ladies,\newline \textbf{mother},\newline \textbf{mothers},\newline \textbf{daughter},\newline \textbf{daughters},\newline she,\newline her,\newline hers
         \end{tabular}
        \vspace*{-6.2ex}
     \end{subfigure}

     \vspace*{3mm}
 \caption{The gender-based terms that we use for Visual Genome and MS-COCO to operationalize binary \gender. This list comes from \cite{https://doi.org/10.48550/arxiv.2301.11100}. Following that work, we exclude images annotated with terms that correspond to \texttt{person} or \texttt{people}. In our \texttt{Evaluation V3}, we experiment with removing the terms that are frequently used with images of animals, shown in \textbf{Bold}.}
\label{tab:gendered}
 \end{table*}

\subsubsection{Construction of confidence intervals.}
\label{app:cs1_ci}

We construct confidence intervals over 250 bootstraps (sampling with replacement each time), then showing 95-percentile error bars.
For \textit{aggregate measurements}, we calculate the mean of the metric for each concept for a given bootstrap.
We then subtract the mean of the metric for each group for that bootstrap and calculate the mean and confidence intervals for the difference in metric across all bootstraps.
For \textit{concept-level measurements}, we take the mean and confidence intervals of the difference of the metric between groups for each concept.

\subsubsection{Additional Results}
\label{app:cs1_res}

\paragraph{Concepts with greatest variation across evaluations with the Visual Genome.}

\begin{figure}[!htb]
     \begin{subfigure}[b]{0.95\textwidth}
         \includegraphics[width=\textwidth]{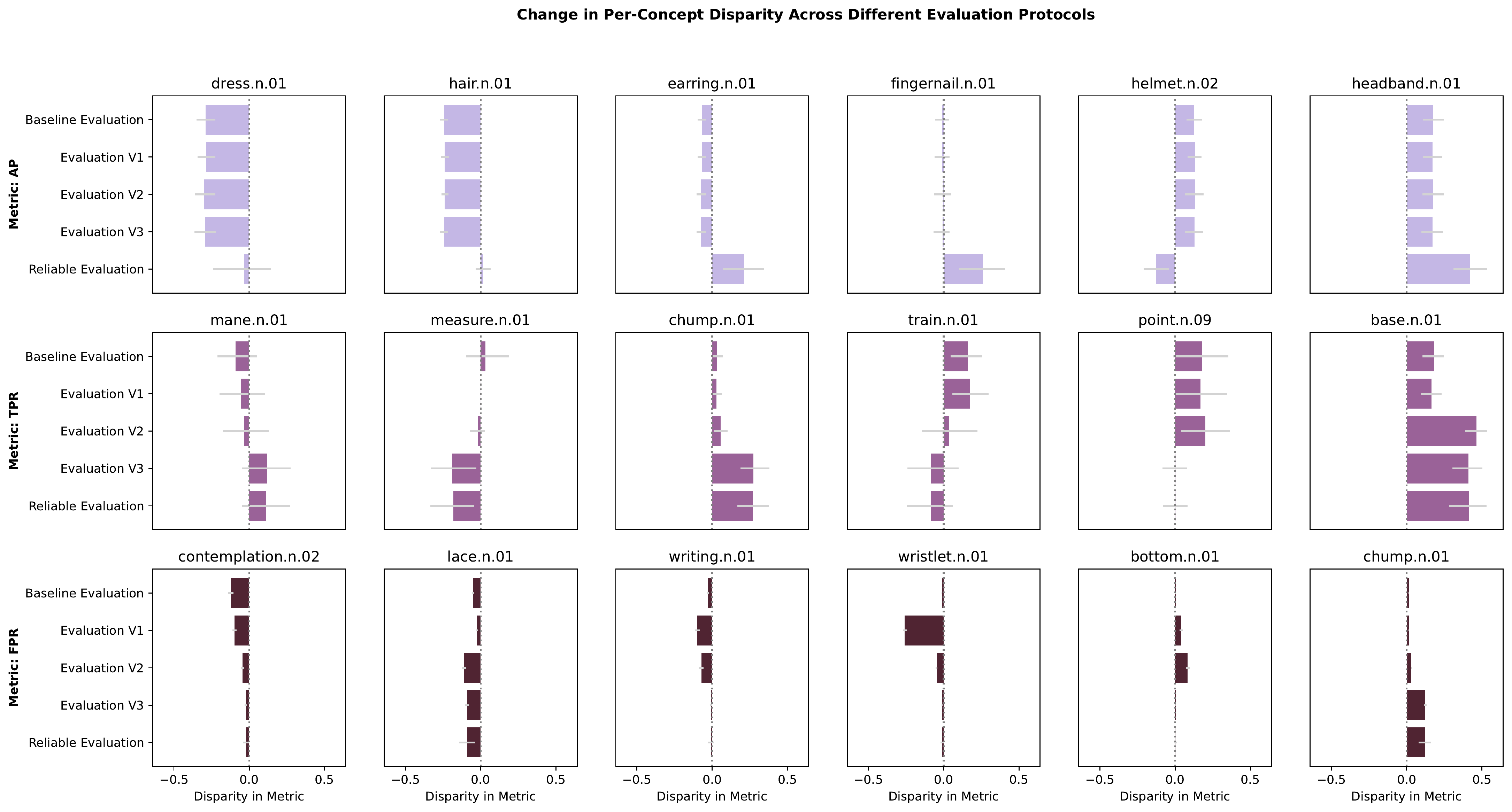}
     \end{subfigure}
     \caption{
     The operationalization of groups and and method of sampling affects the direction and magnitude of per-concept disparities of CLIP when evaluated with the Visual Genome.
               Disparities are shown as the differences between \gender groups, where a positive disparity means that the model has a higher value for the ``man'' \gender group than the ``woman'' \gender group.
     The error bars are 95\% percentile intervals constructed with 250 bootstraps.
     }
     \label{fig:case1_path_eval_alt}
\end{figure}

Figure \ref{fig:case1_path_eval_alt} shows concepts with greatest variation across evaluations with the Visual Genome.

\paragraph{Disparity measurements aggregated across concepts.}

\begin{figure}
     \centering
     \begin{subfigure}[b]{0.75\textwidth}
         \centering
         \includegraphics[width=\textwidth]{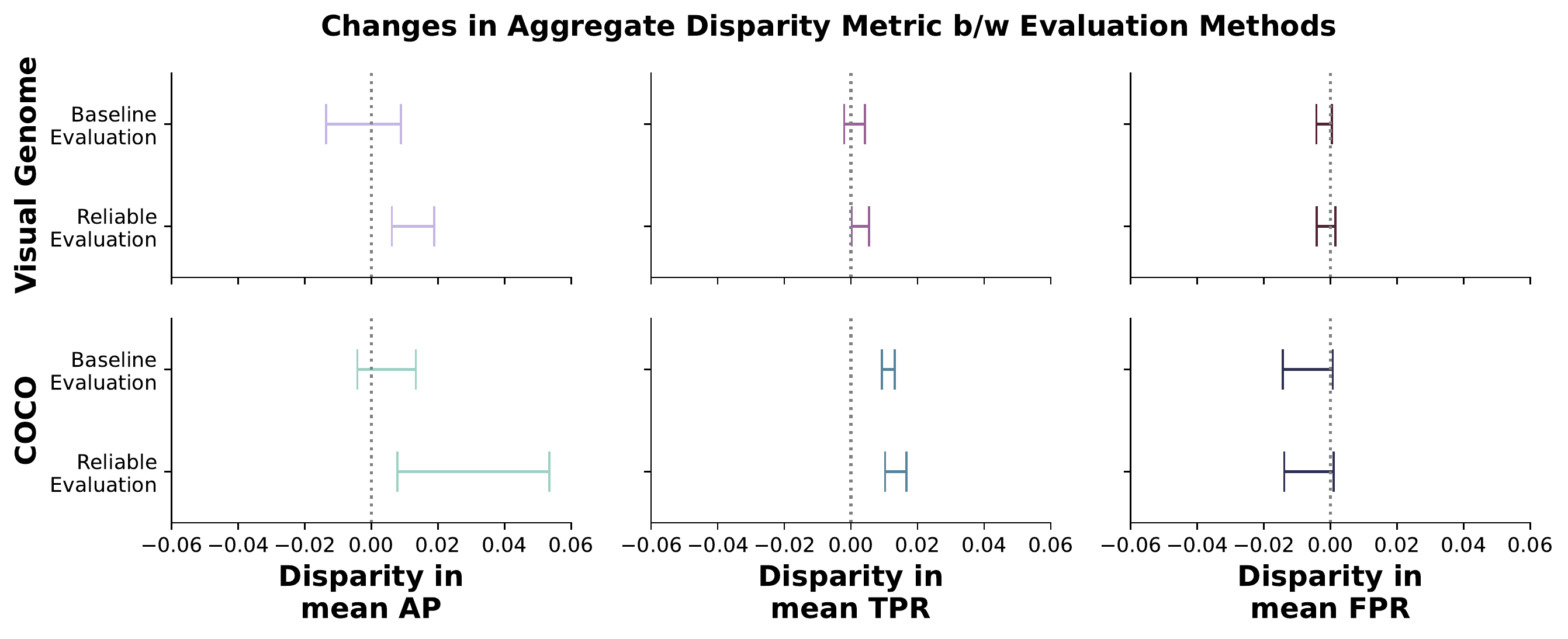}
     \end{subfigure}
     \hfill
     \caption{
     Different choices when evaluating the overall performance of a model can alter observed disparities when using AP.
     Disparities are shown as the differences between \gender groups, where a positive disparity means that the model has a higher value for the ``man'' \gender group than the ``woman'' \gender group.
     The error bars are 95\% percentile intervals constructed with 250 bootstraps.}
\label{fig:case1_path_eval_agg}
\end{figure}

Figure \ref{fig:case1_path_eval_agg} shows how different choices in evaluation can affect analyses when examining disparities over all concepts more when using average precision than TPR/FPR.
This pattern is consistent across both the Visual Genome and MS-COCO datasets.

\begin{figure}[!h]
     \centering
     \begin{subfigure}[b]{0.95\textwidth}
         \centering
         \includegraphics[width=\textwidth]{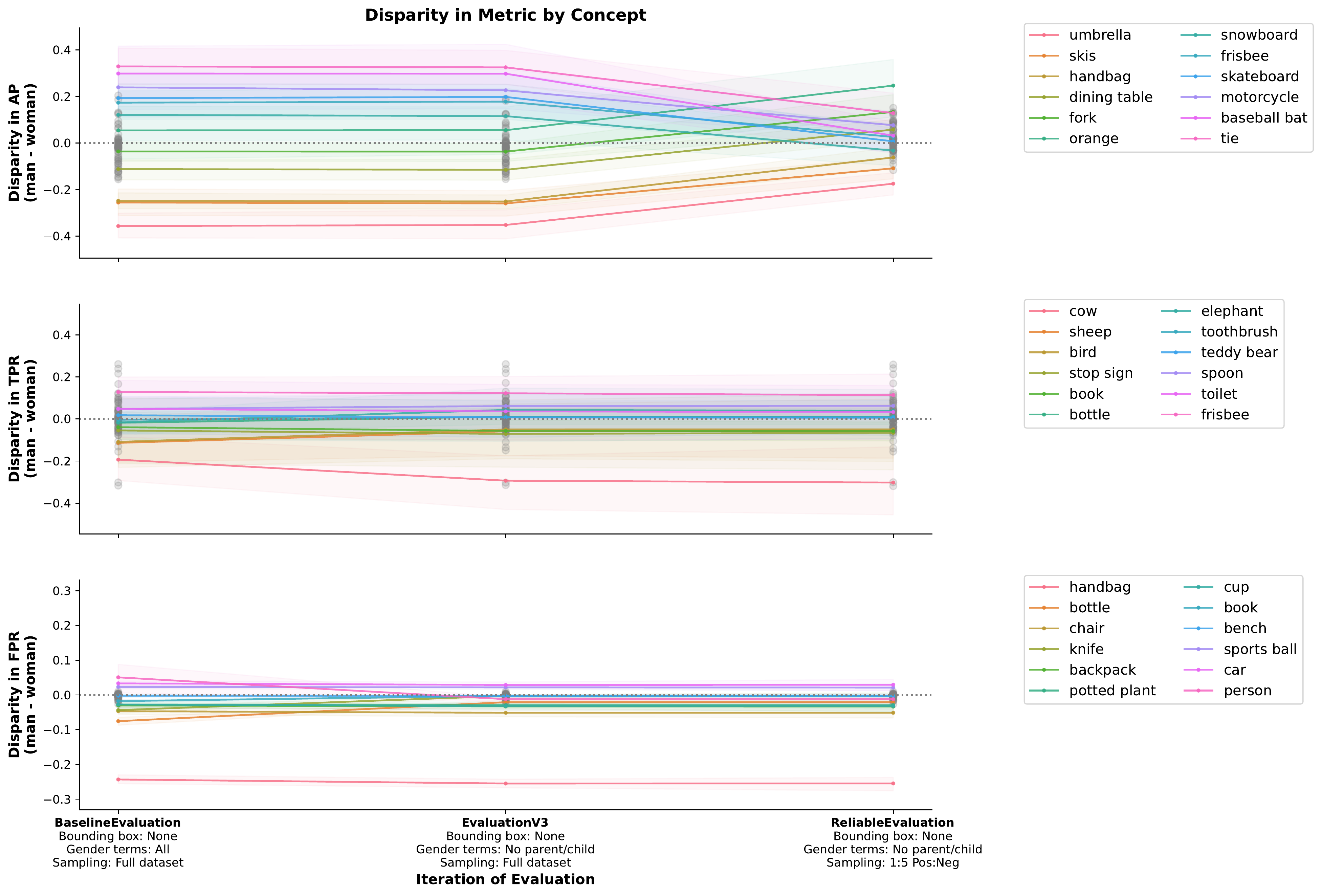}
     \end{subfigure}
     \caption{
     Per-concept disparity measurements of CLIP performance when evaluated on COCO are shown over different \texttt{Decisions} of evaluation.
     Disparities are shown as the differences between \gender groups, where a positive disparity means that the model has a higher value for the ``man'' \gender group than the ``woman'' \gender group.
     The shaded error bars are 95\% percentile intervals constructed with 250 bootstraps.
     }
     \label{fig:case1_path_eval_coco}
\end{figure}

\paragraph{Per-concept disparity evaluations using COCO}
In Figure \ref{fig:case1_path_eval_coco} we show per-concept disparity metrics for the MS-COCO \citep{lin2014microsoft} dataset.
Because we use only captions for determining groups, and not bounding boxes, we skip \texttt{Evaluation V1} and \texttt{Evaluation V2} which focused on filtering the bounding box used for determining groups.

We observe that, as with Visual Genome, the method of sampling affects average precision much more than TPR and FPR.
However, compared to the Visual Genome, there is a relatively small effect of removing the gender-based terms corresponding to parent and child relationships.
This suggests that this annotation pattern associating animals with parental relationships may not occur often in MS-COCO, and could be specific to the Visual Genome.

\begin{figure}
     \centering
     \begin{subfigure}[b]{0.95\textwidth}
         \centering
         \includegraphics[width=\textwidth]{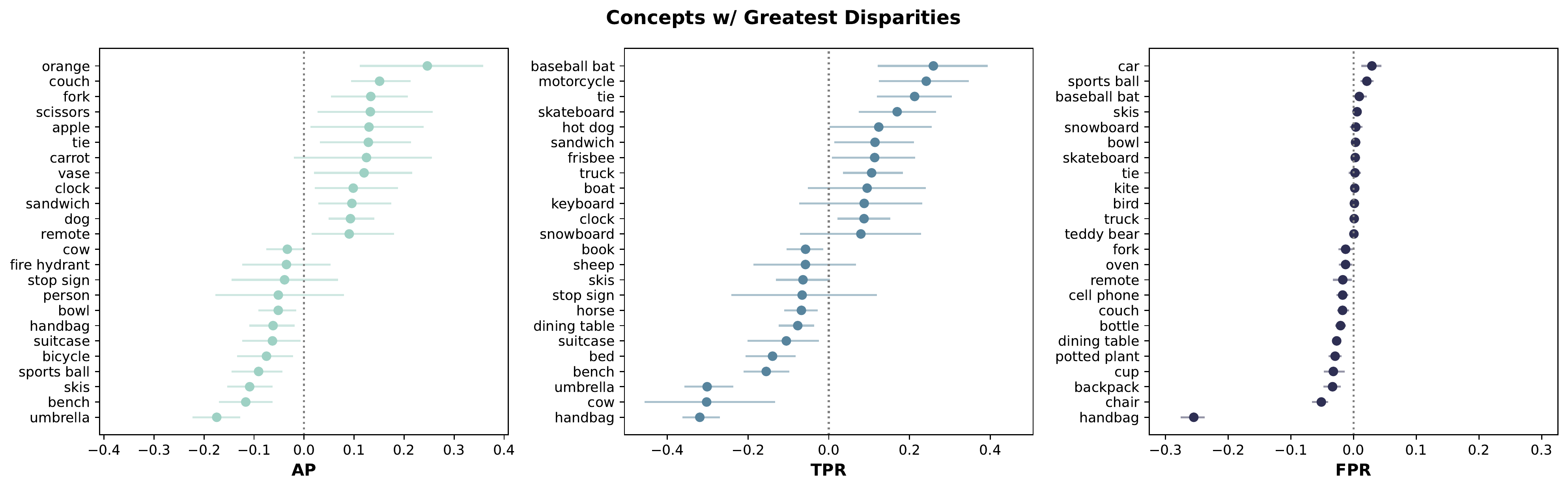}
     \end{subfigure}
     \hfill
     \caption{
     Concepts with the greatest disparities in performance using the \finaleval Method. Disparities are shown as the differences between \gender groups, where a positive disparity means that the model has a higher value for the ``man'' \gender group than the ``woman'' \gender group.
     The error bars are 95\% percentile intervals constructed with 250 bootstraps.}
\label{fig:case1_greatest_disp_coco}
\end{figure}

As with the Visual Genome, we see that the choice of sampling has a much larger effect on average precision measurements than TPR and FPR measurements.
In Figure \ref{fig:case1_greatest_disp_coco} we show the concepts with the greatest disparities using the \finaleval method.
These concepts are included in our qualitative analysis of large disparities presented in Section \ref{sec:root_cause}.

\subsection{Additional Details: Case Study \#2}

\subsubsection{Comparing the two class mappings.}
\label{app:cs2_mappings}
Examples of how the DollarStreet labels are translated to different ImageNet classes between
the two mappings are shown in Figure \ref{fig:class_mapping}.
We keep the original DollarStreet labels fixed; because some of the DollarStreet labels in the alternative ImageNet-1K mapping that we used are not present in our original DollarStreet dataset, we omit those ImageNet-1K classes.

\begin{figure}
	\begin{minipage}{0.9\linewidth}
		\centering
		\begin{tabular}[width=\linewidth]{p{2.5cm}  p{5.75cm}  p{3.3cm} }
             \toprule
              DollarStreet Label & ImageNet22K Class & ImageNet1K Class
              \\ \midrule
                  phones  &
                  telephone, phone, telephone\_set  &
                  cellphone
                \\ \midrule
                  roofs  &
                  roof &
                  tile roof
                \\ \midrule
                  bikes  &
                  bicycle, bike, wheel, cycle &
                  all-terrain bike
                 \\ \midrule
                  parking lots  &
                  garage &
                  parking meter
                  \\ \midrule
                  lock on front door &
                  lock &
                  padlock
              \\ \bottomrule
            \end{tabular}
	\end{minipage}\hfill
\caption{Examples of mappings between DollarStreet labels and model classes, using mappings that are defined in previous works.}
\label{fig:class_mapping}
\end{figure}

We report per-concept results based on the ImageNet classes, not the DollarStreet classes, as this maps more directly to model behavior.
This means that we may have an image has a DollarStreet label "showers" and multiple ImageNet labels: "shower\_room", "shower", "bathtub, bathing\_tub, bat, tub."
When calculating hit-rate, the image is considered as a "hit" if any one of the ImageNet classes are in the top-5.
When calculating AP, we evaluate each ImageNet concept distinctly and include as negative images any image that does not include the ImageNet concept.

\subsubsection{Rare concepts can have large disparities.}
\label{app:cs2_rare}
\begin{figure}[!h]
     \centering
     \begin{subfigure}[b]{0.60\textwidth}
         \centering
         \includegraphics[width=\textwidth]{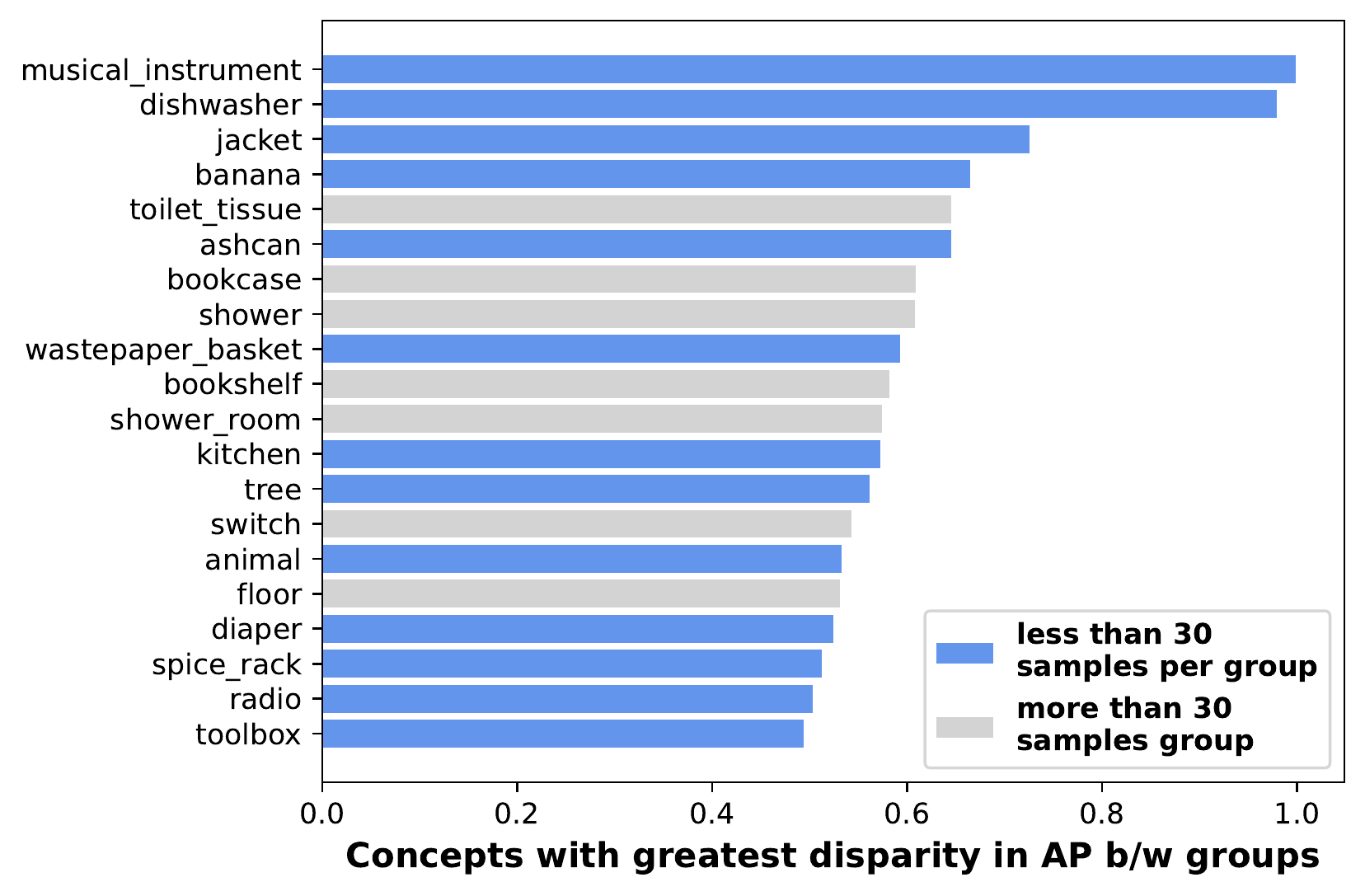}
     \end{subfigure}
     \caption{
     Many concepts with the largest disparities in DollarStreet when evaluating with AP have fewer than 30 samples per group.
     }
     \label{fig:case2_all_samples}
\end{figure}

In Figure \ref{fig:case2_all_samples} we show the concepts in DollarStreet with the largest disparity in AP prior to filtering out concepts with few samples.
Many concepts have fewer than 30 samples per group and can be susceptible to noise.
For example, the concept \texttt{musical\_instrument} has only one sample for Africa and seven samples for Europe, and the concept \texttt{dishwasher} has only two samples for Africa and four samples for Asia.
We use the ImageNet22K mapping for these measurements, and the disparities shown are the maximum difference in AP between pairs of continents.

\subsubsection{Construction of confidence intervals.}
\label{app:cs2_ci}
We construct confidence intervals over 250 bootstraps (sampling with replacement each time), then showing 95-percentile error bars.
For \textit{aggregate measurements}, we calculate the mean of the metric for each concept for a given bootstrap.
We then subtract the mean of the metric for each group for that bootstrap and calculate the mean and confidence intervals for the difference in metric across all bootstraps.
For \textit{concept-level measurements}, we take the mean and confidence intervals of the difference of the metric between groups for each concept.

\subsection{Additional Details: Root Cause Analyses}
\label{app:qual}

\begin{figure}
        \begin{tabular}{p{15cm}}
        \small{\texttt{dress.n.01} (Visual Genome)}
         \small{\texttt{umbrella} (COCO)}
        \tab \tab \tab \tab \tab \tab \small{\texttt{cow} (COCO)}
        \\
        \includegraphics[height=18mm]{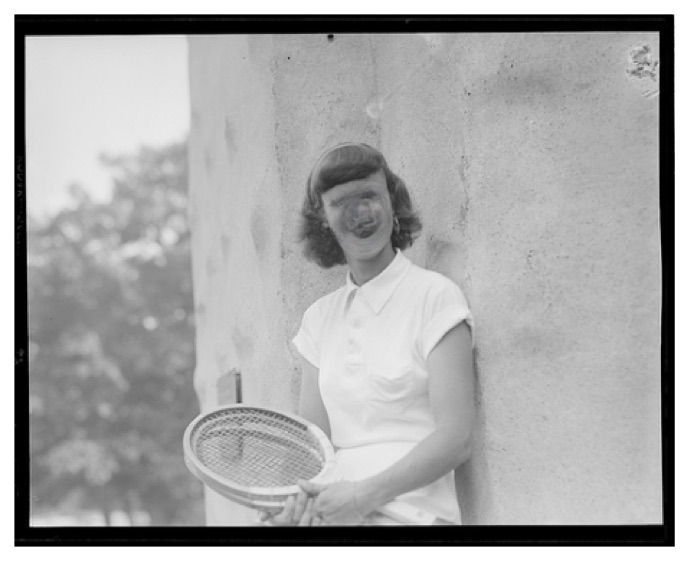}
        \includegraphics[height=18mm]{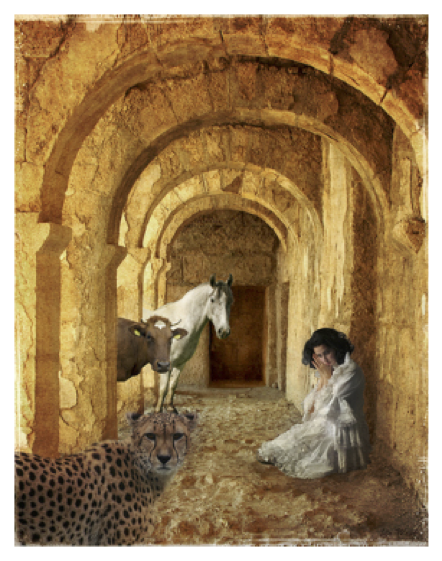}
        \tab
        \includegraphics[height=18mm]{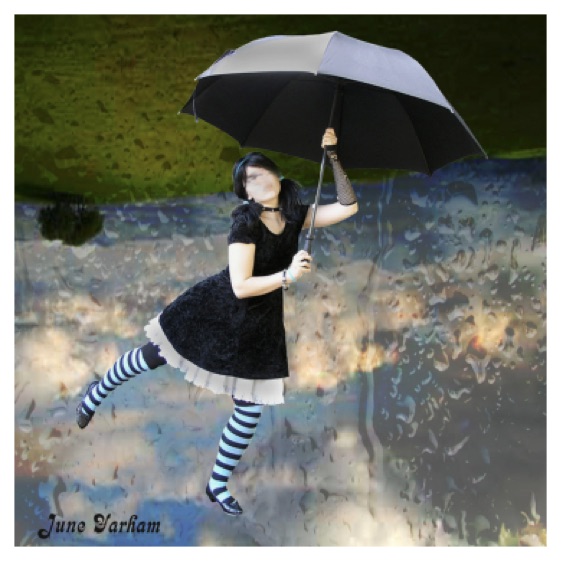}
        \includegraphics[height=18mm]{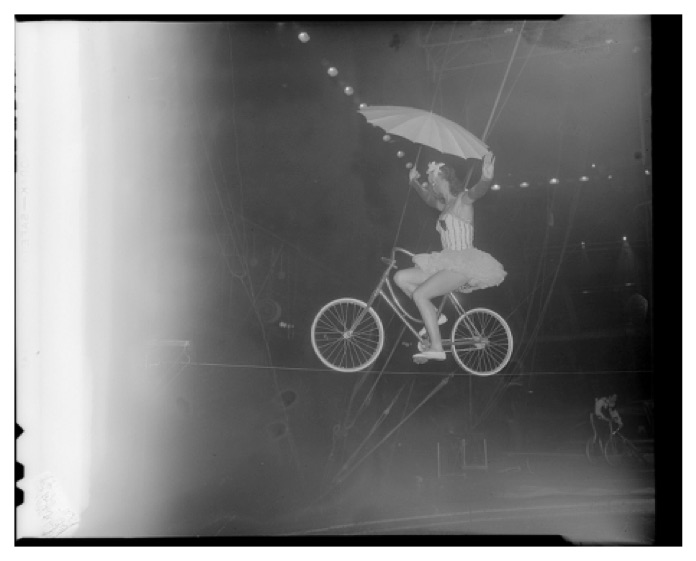}
        \tab
        \includegraphics[height=18mm]{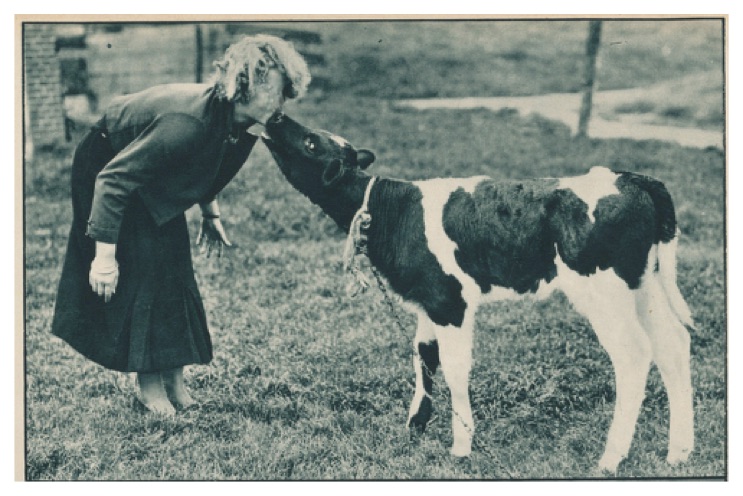}
        \includegraphics[height=18mm]{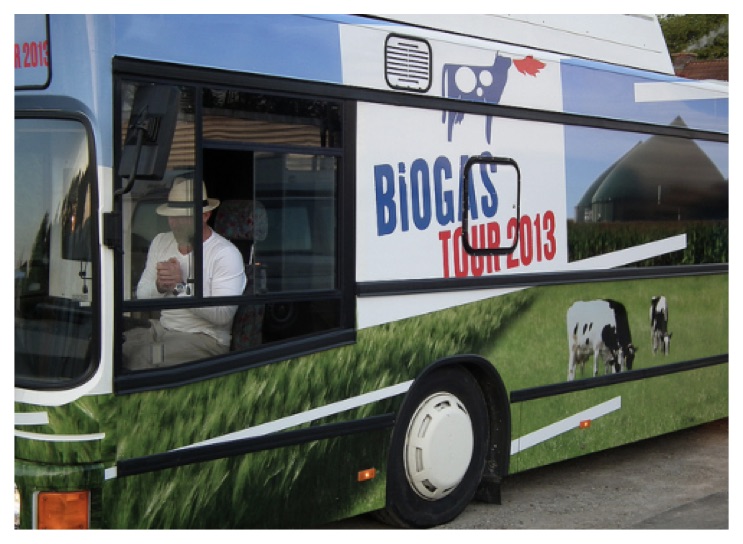}
  \end{tabular}
\caption{Concepts with large disparities contain images depicting different media. We blur faces for the privacy of those depicted and use the original images in our evaluations.}
\label{fig:medium}
\end{figure}

Figure \ref{fig:medium} shows examples of how the Visual Genome and COCO have images that depict different media rather than a standard photograph, such as art pieces, black and white photographs, or signs.
These tend to contain women more than men.

\end{document}